\RequirePackage{amsmath}
\documentclass{svjour3}
\usepackage{natbib}
\usepackage{amssymb}
\usepackage{amsmath}
\usepackage{tabularx}
\usepackage{caption}
\usepackage{subcaption} %for 2 figures side by side
\usepackage{graphicx}
\usepackage{epstopdf}      %for eps graphics
\DeclareGraphicsExtensions{.eps}
\usepackage[makeroom]{cancel}
\usepackage{array}
\usepackage{multirow}
\usepackage{placeins}
\usepackage{float}

\newcolumntype{?}{!{\vrule width 2pt}}
\def\argmin{\mathop{\rm argmin}}
\def\argmax{\mathop{\rm argmax}}

\usepackage{tablefootnote}

\begin{document}
	\raggedbottom
	\title{Automated Adaptation Strategies for Stream Learning}
	
	\author{Rashid~Bakirov \and
		Damien~Fay  \and
		Bogdan~Gabrys
	}
	% <-this % stops a space
	\institute{R. Bakirov \at
		Department of Computing and Informatics, Bournemouth University, Poole, UK.
		\email{rbakirov@bournemouth.ac.uk}  
		\and
		D. Fay \at INFOR/Logicblox, Atlanta, Georgia, US
		\and
		B. Gabrys \at
		Advanced Analytics Institute, University of Technology Sydney, Australia. 
	}

	\maketitle
	
	\begin{abstract}
		Automation of machine learning model development is increasingly becoming an established research area. While automated model selection and automated data pre-processing have been studied in depth, there is, however, a gap concerning automated model adaptation strategies when multiple strategies are available. Manually developing an adaptation strategy can be time consuming and costly. In this paper we address this issue by proposing the use of flexible adaptive mechanism deployment for automated development of adaptation strategies. Experimental results after using the proposed strategies with five adaptive algorithms on 36 datasets confirm their viability. These strategies achieve better or comparable performance to the custom adaptation strategies and the repeated deployment of any single adaptive mechanism. 
	\end{abstract}

	% \fi

	\section{Introduction}
	\label{sec:intro}
	
	Automated model selection has long been studied \citep{Wasserman2000} and recently, notable advances in practical automated machine learning (AutoML) approaches \citep{Hutter2011,Lloyd2014,Kotthoff2017,Mohr2018,MartinSalvador2019, Olson2019,kedziora2020autonoml} have been made. In addition, automated data pre-processing in the context of complex machine learning pipelines generation and validation has also been a topic of recent interest \citep{Feurer2015,MartinSalvador2019,Nguyen2020}. There is however a gap concerning automated development of models' adaptation strategy, which is addressed in this paper. Here we define \textit{adaptation} as changes in model training set, parameters and structure, all designed to track changes in the underlying data generating process over time. This contrasts with model selection which focuses on parameter estimation and the family to sample the model from.
	
	With the current advances in data storage, database and data transmission technologies, learning on streaming data has become a critical part of many processes. Many models which are used to make predictions on streaming data are static, in the sense that they do not learn on current data and hence remain unchanged. However, there exists a class of models, stream learning models, which are capable of adding observations from the data stream to their training sets. In spite of the fact that these models utilise the data as it arrives,  there can still arise situations where the underlying assumptions of the model no longer hold. We call such settings dynamic environments, where changes in data distribution \citep{Zliobaite2011}, change in features' relevance \citep{Fern2000}, non-symmetrical noise levels \citep{Schmidt2007} are common. These phenomena are sometimes called \textit{concept drift}. It has been shown that many changes in the environment which are no longer being reflected in the model contribute to the deterioration of model's accuracy over time \citep{Schlimmer1986,streetkim2001,Klinkenberg2004,coltermaloof2007}. This requires constant manual retraining and readjustment of the models which is often expensive, time consuming and in some cases impossible - for example when the historical data is not available any more. Various approaches have been proposed to tackle this issue by making the model adapt itself to the possible changes in environment while avoiding its complete retraining. These approaches however are manually designed and the application of automated machine learning to streaming data is scarce, which is the gap we aim to contribute to.
	
	Typically there are several possible ways or \textit{adaptive mechanisms} (AMs) to adapt a given model. A single iteration of adaptation is achieved by deploying one of multiple AMs (including trivial "doing nothing"), which changes the state of the existing model. Thus, during the model's operation, it is adapted by the sequential deployment of various AMs with the arrival of new data. We call the order of this deployment an \textit{adaptation strategy} (AS). While in most of the existing research these adaptation strategies are custom (i.e. algorithm-specific) and are fixed at the design stage of the algorithm, a sequential adaptation framework proposed in our earlier work \citep{Bakirov2015} enables flexible adaptation strategies without a prescribed AM deployment order. These flexible adaptation strategies, automatically developed according to this framework can be applied to any set of adaptive mechanisms for various machine learning algorithms. This removes the need to design custom adaptive strategies, resulting in automation of adaptation process. In this work we empirically show the viability of the automated adaptation strategies based on cross-validation \citep{Bakirov2015} with the optional use of retrospective model correction \citep{Bakirov2016}.  
	
	We focus on the batch prediction scenario, where data arrives in large segments called batches. This is a common industrial scenario, especially in the chemical, microelectronics and pharmaceutical areas \citep{Cinar2003}. For the experiments we use Simple Adaptive Batch Learning Ensemble (SABLE) \citep{Bakirov2015} and batch versions of four popular stream learning algorithms - the Dynamic Weighted Majority (DWM) \citep{coltermaloof2007}, the Paired Learner (PL) \citep{Bach2010}, the Leveraged Bagging (LB) \citep{Bifet2010} and BLAST \citep{Rijn2015having}. The use of these five algorithms allows to explore different types of online learning methods; local experts ensemble for regression in SABLE, global experts ensemble for classification in DWM and LB, switching between the two models in PL and the heterogeneous global ensemble in BLAST.
	
	After a large-scale experimentation with 5 regression and 31 classification datasets, the main finding of this work is that in our settings, the proposed automated adaptive strategies show comparable accuracy rates to the custom adaptive strategies and, in many cases, to the repeated deployment of a single ``best" AM. Thus, they are feasible to use for adaptation purposes, while saving time and effort spent on designing custom strategies.
	
	The paper follows by presenting the related work on automated machine learning and adaptive mechanisms in Section \ref{sec:relatedworks}. Section \ref{section:formulation} presents mathematical formulation of the framework of adaptation with multiple adaptive mechanisms in batch streaming scenario. Section \ref{section:algorithm} introduces algorithms used for the experimentation, including their inherent adaptive mechanisms and custom adaptation strategies. Experimental methodology, the datasets on which experiments were performed and results are given in Section \ref{section:experiments}. We give our final remarks in Section \ref{section:conclusions}.
	
	\section{Related Work}
	\label{sec:relatedworks}
	
	This section provides a background for our research. We start with a review of relevant automated machine learning approaches, particularly those which consider streaming data scenario. We follow up with a broad analysis of ML literature from the  adaptive mechanisms point of view, where we introduce a simple hierarchy of adaptation. We then discuss how multiple adaptive mechanisms paradigm has been used for automating the design of predictive algorithms.
	
	\subsection{Automated machine learning for streaming data}
	\label{sec:automl}
	Automated machine learning is an active research area. So far however, it has been mostly applied to static datasets, and there are not many works which consider automation for streaming scenario. Among these, different approaches exist. One of the works, before the most recent wave of AutoML research, can be found in \citep{Kadlec2009} where a general purpose architecture to develop robust, adaptive prediction systems for the autonomous operation in changing environments for streaming data has been proposed. Various instantiations of this architecture followed focusing on challenging problems from the process industry when building adaptive, predictive soft sensors \citep{Kadlec2010,Kadlec2011a,Bakirov2017}. 
	
	Taking advantage of the recent wave of research in AutoML, an alternative approach to adaptation to changing environments was proposed in \citep{MartinSalvador2016} where repeated automated deployment of Auto-WEKA for Multi-Component Predictive Systems (MCPS) to learn from new batches of data was used for life-long learning and the adaptation of complex MCPS when applied to changing streaming data from process industries. \cite{Celik2020} represent a development of this idea with the inclusion of the drift detection and the experimentation using several open source AutoML frameworks. 
	An interesting approach closely tied with the Auto\_Sklearn is described in \citep{Madrid2019}. Authors propose using the ensemble nature of this framework to deal with streaming data, by adapting the weights of experts and adding new ones. 
	
	Some of the other recently proposed relevant methods are primarily focused on hyper-parameter optimisation problems. For example, \cite{Veloso2018} propose hyper-parameter optimization for streaming regression problems using the Nelder-Mead algorithm. In their experiments they optimise the hyper-parameters of one specific regression method. \cite{Carnein2020} are focusing on the hyper-parameter selection for clustering of data in a streaming environment. They propose utilising a dynamic ensemble of different hyper-parameter configurations. 
	
	Despite the existing research, as acknowledged and discussed in a recent  comprehensive and synthesising review of concepts in AutoML research and beyond \citep{kedziora2020autonoml}, the pursuit of autonomy, described as the AutoML system's capability to independently adapt the ML solution over a lifetime of operation in changing environments, remains a lofty goal.
	
	\subsection{Adaptive mechanisms}
	\label{sec:mam}
\begin{figure}[t]
		\begin{center}
			\includegraphics[scale=0.5]{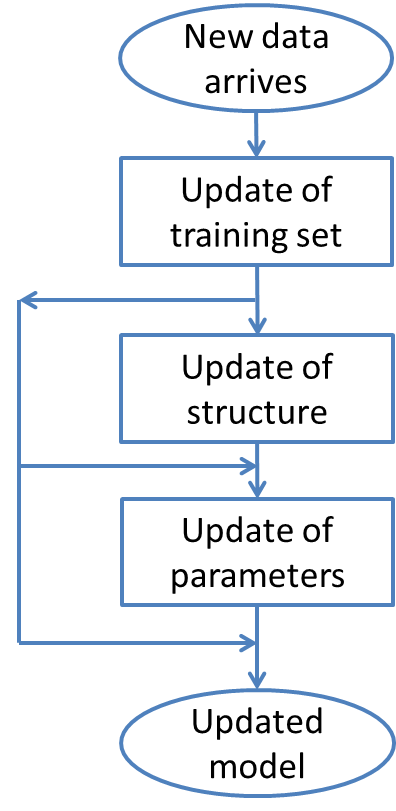}
			\caption{General adaptation scheme \citep{Bakirov2017a}.}
			\label{fig:adapt}
		\end{center}
	\end{figure}

	Adapting machine learning models is an essential strategy for automatically dealing with changes in an underlying data distribution to avoid training a new model manually. Modern machine learning methods typically contain a complex set of elements allowing many possible AMs. This can increase the flexibility of such methods and broaden their applicability to various settings. However, the existence of multiple AMs also increases the decision space with regards to the adaptation choices and parameters, ultimately increasing the complexity of adaptation strategy. 
	A possible hierarchy\footnote{Here, the hierarchy is meant in a sense that the application of an adaptive mechanism of the higher level, requires the application of the adaptive mechanism of lower level.} of AMs is presented in Figure \ref{fig:adapt}.
	
	In a streaming data setting, to increase the accuracy, it can be beneficial to include recent data in the training set of the predictive models. On the other hand however, retraining a model from scratch is often inefficient, particularly dealing with high throughput scenarios or even impossible when the historical data is no longer available. For these cases, the solution is updating the model using only the available recent data. This can be done inherently by some general purpose ML algorithms, e.g. Naive Bayes or using stream/online algorithms, e.g. online Least Squares Estimation \citep{Jang1997}, online boosting and bagging \citep{Oza2001} etc. Additionally, for non-stationary data, it becomes important to not only select a training set of sufficient size but also one which is relevant to the current data. This is often achieved by a moving window \citep{Widmer1996,Klinkenberg2004,Zliobaite2010} or decay approaches\citep{JoeQin1998,Klinkenberg2000}. 
	
	The structure of a predictive model is a graph with the set of its components and the connections therein. Some common examples are hierarchical models (e.g. decision trees) or more complex graphs (e.g. Bayesian
	or neural networks). Here, the structure is not necessarily limited to the topological context -- number of rules in rule based systems or number of experts in an ensemble could be considered part of the model's structure. Adaptation can be achieved by updating 
	this structure, for example in decision and model trees \citep{Domingos2000,Hulten2001,Ikonomovska2010}, neuro-fuzzy approaches \citep{Gabrys1999, Gabrys2004, Sahel2007}, neural networks \citep{Carpenter1991,Vakil-Baghmisheh2003,Ba2013}, Bayesian networks \citep{Friedman1997,Alcobe,Castillo2006} and ensemble methods \citep{Stanley2002,Gabrys2006,coltermaloof2007,Hazan2009,Lemke2009,Ruta2011,GomesSoares2015,Bakirov2017}.
	
	The final layer of adaptation is changing the models' parameters, e.g. experts' combination weights in ensemble methods. These weights are often recalculated or updated throughout a models' runtime \citep{Littlestone1994,coltermaloof2007,Elwell2011,Kadlec2011a,Bakirov2017}. Another group of techniques belonging to this family are methods using meta-learning for model adaptation \citep{Nguyen2012heterogeneous,Rossi2014meta, Rijn2015having,Lemke2010}. These methods generally include training a meta-model using meta-features. The meta-model is then used to select one or more predictors to calculate the final prediction. The change of the meta-model can then be seen as the change in parameters of the predictive model.
	
	In this work we consider the possibility of using multiple different adaptive mechanisms, most often at different levels of the hierarchy. Many modern machine learning algorithms for streaming data explicitly include this possibility. A prominent example are the adaptive ensemble methods \citep{Wang2003,coltermaloof2007,Scholz2007,Bifet2009,Kadlec2010,Elwell2011,Alippi2012,Souza2014,GomesSoares2015,Bakirov2017} which often feature AMs from all three levels of hierarchy - online update of experts, changing experts' combination weights and modification of experts' set. Machine learning methods with multiple AMs are not limited to ensembles, but can also include Bayesian networks \citep{Castillo2006}, decision trees \citep{Hulten2001}, model trees \citep{Ikonomovska2010}, champion-challenger schemes \citep{Bach2010} etc.
	
	\subsection{Automating design of algorithms with multiple AMs}
	\label{sec:automlwithmam}
	Existence of multiple AMs raises questions w.r.t. how they should be deployed. This includes defining the order of deployment and adaptation parameters (e.g. decay factors, expert weight decrease factors, etc.). It should be noted that all of the aforementioned algorithms use custom adaptive strategies, meaning that they deploy AMs in a manner specific to each of them. It follows that designing adaptive machine learning methods is a complex enterprise and is an obstacle to the automation of machine learning model's design. \cite{Kadlec2009} present a plug and play architecture for pre-processing, adaptation and prediction which foresees the possibility of using different adaptation methods in a modular fashion, but does not address the method of AM selection. \cite{Bakirov2015,Bakirov2016} have presented several such methods for AM selection for their adaptive algorithm, which are discussed in detail in Section \ref{sec:strategies}. These methods can be seen as automated adaptive strategies, which are applicable to all adaptive machine learning methods with multiple AMs. This allows simply using the described strategies for model adaptation, once having defined the available AMs.
	
	\section{Formulation}\label{section:formulation}
	
	% \subsection{General formulation}
	
	As adaptation mechanisms can affect several elements of a model and can depend on performance several time steps back, it is necessary to clarify the concepts via a framework to avoid confusion. We assume that the data is generated by an unknown {time varying} data generating process which can be formulated as:
	\begin{equation}
		y_\tau=\psi(\boldsymbol{x}_{{\tau}},\tau)+\epsilon_\tau,
		\label{e:eq1}
	\end{equation}
	where $\psi$ is the unknown function, $\epsilon _\tau$ a noise term, \mbox{$\boldsymbol{x}_\tau \in \mathcal{R}^M$} is an input data instance, and $y_\tau$ is the observed output at time $\tau$. Then we consider the predictive method at a time $\tau$ as a function:  
	\begin{equation}
		\hat{y}_\tau=f_\tau(\boldsymbol{x}_{{\tau}},\Theta_f),
		\label{e:eq2}
	\end{equation}
	where $\hat{y}_\tau$ is the prediction, $f_\tau$ is an approximation (i.e. the model) of $\psi(\boldsymbol{x},\tau)$, and $\Theta_f$ is the associated parameter set. Our estimate, $f_\tau$, evolves via adaptation as each batch of data arrives as is now explained.  
	
	\subsection{Adaptation}
	\label{sec:adaptation}
	In the batch streaming scenario considered in this paper, data arrives in batches with \mbox{$\tau\in \{\tau_{k}\cdots \tau_{k+1}-1\}$}, where $\tau_k$ is the start time of the $k$-th batch. If $n_k$ is the size of the $k$-th batch, $\tau_{k+1}=\tau_{k}+n_k$. It then becomes more convenient to index the model by the batch number $k$, denoting the inputs as $\boldsymbol{X}_k =\boldsymbol{x}_{\tau_{k}},\cdots, {\boldsymbol{x}}_{\tau_{k+1}-1}$ and the outputs as $\boldsymbol{y}_k = y_{\tau_{k}},\cdots, {y}_{\tau_{k+1}-1}$. We examine the case where the prediction function $f_k$ is static within a $k$-th batch.\footnote{A batch typically represents a meaningful real-world segmentation of the data, for example a plant run and so our adaptation attempts to track run to run changes in the process.}
	
	We denote the \textit{a priori} predictive function at batch $k$ as $f_k ^{-}$, and the \textit{a posteriori} predictive function, i.e. the adapted function given the observed output, as $f_k ^{+}$. An \textit{adaptive mechanism}, $g(\cdotp)$, may thus formally be defined as an operator which generates an updated prediction function based on the batch ${\boldsymbol{\mathcal{V}}}_k=\{\boldsymbol{X}_k, \boldsymbol{y}_k\}$ and other optional inputs. This can be written as:
	\begin{equation}
		g_k(\boldsymbol{X}_k, \boldsymbol{y}_k, \Theta_g, f_k^{-}, \hat{\boldsymbol{y}}_k):f_k^{-} \rightarrow f_{k}^{+}. 
	\end{equation}
	or alternatively as $f_k^{+} = f_k^{-} \circ g_k$ for conciseness. Note $f_k^{-}$ and $\hat{\boldsymbol{y}}_k$ are optional arguments and $\Theta_g$ is the set of parameters of $g$. The function is propagated into the next batch as $f_{k+1}^{-} = f_{k}^{+}$ and predictions themselves are always made using the \textit{a priori} function $f_{k}^{-}$.
	
\begin{figure*}[]
		\centering
		\begin{subfigure}[b]{0.5\linewidth}
			\centering\includegraphics[width=\linewidth]{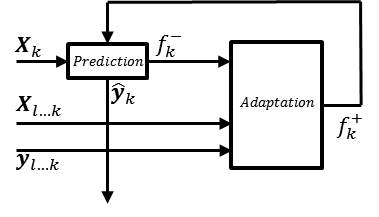}
			\caption{}
			\label{fig:scheme}%
		\end{subfigure}%
		\begin{subfigure}[b]{0.5\linewidth}
			\centering\includegraphics[width=\linewidth]{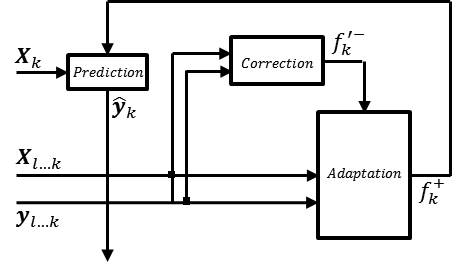}
			\caption{}
			\label{fig:scheme2}%
		\end{subfigure}
		\caption{(a) Adaptation scheme. (b) Adaptation scheme with retrospective correction. Here $1\leq l \leq k$ and $f_k^{'-}$ represents the result of retrospective correction. Depending on the algorithm, inputs can be optional.}
		
	\end{figure*}

	We examine a situation when a choice of multiple, different AMs,\\ \mbox{$\{\emptyset, g_1,...,g_H\}=G$}, is available. Any AM $g_{{h}_k} \subset G$ can be deployed on each batch, where ${h_k}$ denotes the AM deployed at batch $k$.  As the history of all adaptations up to the current batch, $k$, have in essence created $f_k^{-}$, we call that sequence $g_{{h}_1},...,g_{{h}_k}$ an \textit{adaptation sequence}.  Note that we also include the option of applying no adaptation denoted by $\emptyset$. In this formulation, only one element of $G$ is applied for each batch of data. Deploying multiple adaptation mechanisms on the same batch are accounted for with their own symbol in $G$. Figure \ref{fig:scheme} illustrates our initial formulation of adaptation. 
	
	\begin{figure*}[h!]
		\centering
		\begin{subfigure}[c]{0.5\linewidth}
			\centering
			\includegraphics[width=0.7\linewidth]{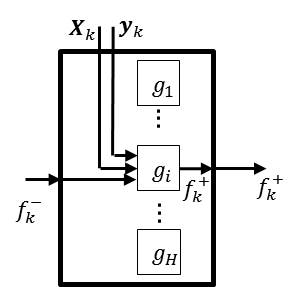}
			\caption{Single}
			\label{fig:simpleadaptscheme}
		\end{subfigure}%
		\begin{subfigure}[c]{0.5\linewidth}
			\centering
			\includegraphics[width=0.8\linewidth]{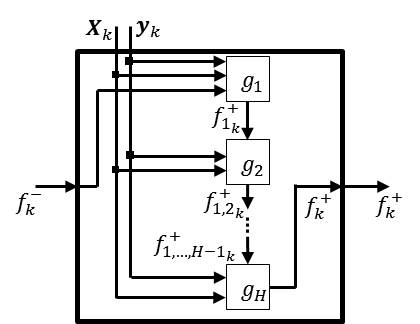}
			\caption{\textit{Multiple}}
			\label{fig:jointadaptscheme}
		\end{subfigure}%
		
		\begin{subfigure}[c]{0.5\linewidth}
			\centering
			\includegraphics[width=\linewidth]{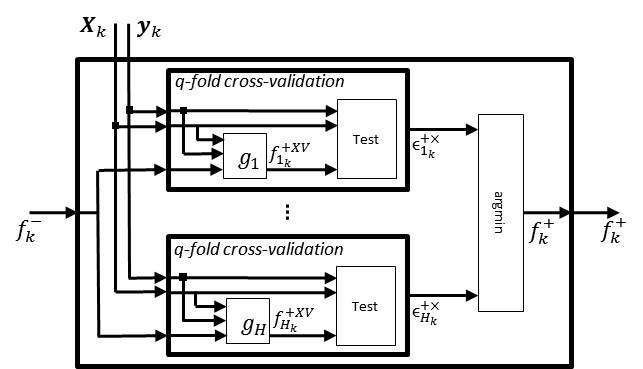}
			\caption{\textsc{XVSelect}}
			\label{fig:xvadaptscheme}
		\end{subfigure}%
		\begin{subfigure}[c]{0.5\linewidth}
			\centering
			\includegraphics[width=\linewidth]{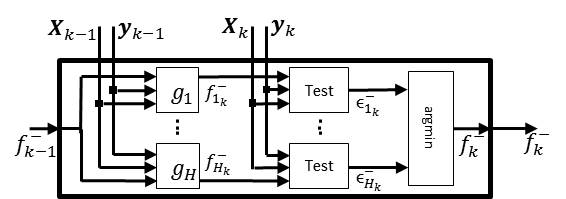}
			\caption{Retrospective correction}
			\label{fig:rcadaptscheme}
		\end{subfigure}%
		
		\caption{Automated adaptation strategies.}
	\end{figure*}
	
	\subsection{Automated adaptation strategies}
	\label{sec:strategies}
	In this section we present different generic automated adaptive strategies offering flexible deployment of AMs, which can be applied to any adaptive algorithm.  
	
	At every batch $k$, an AM $g_{{h}_k}$ must be chosen to deploy on the current batch of data. To obtain a benchmark performance, an adaptation strategy which minimizes the error over the incoming data batch ${\boldsymbol{X}_{k+1}, \boldsymbol{y}_{k+1}}$:
	\begin{equation}
		f_{k+1}^{-}= f_k^{-} \circ g_{{h}_k}, \quad {h}_k= \argmin_{{h}_k \in 1\cdots H} \langle (f_k^{-} \circ g_{{h}_k}) (\boldsymbol{X}_{k+1}), \boldsymbol{y}_{k+1} \rangle
		\label{eq:oracle}
	\end{equation}
	where $\langle \mbox{ } \rangle$ denotes the chosen error measure, can be used. Since ${\boldsymbol{X}_{k+1}, \boldsymbol{y}_{k+1}}$ are not yet obtained, this strategy is not applicable in practice. Also note that this may not be the overall optimal strategy which minimizes the error over the whole dataset. We refer to this strategy as \textsc{Oracle}.  
	
	Given the inability to conduct the \textsc{Oracle} strategy, below we list some alternatives. The simplest adaptation strategy is applying the same AM  to every batch. The scheme of this strategy is given in Figure \ref{fig:simpleadaptscheme}. Note that this scheme fits the ``Adaptation" box in Figure \ref{fig:scheme}. A more common practice (see Section~\ref{sec:relatedworks}) is applying multiple or all available adaptive mechanisms. The scheme of this strategy is given in Figure \ref{fig:jointadaptscheme} which again fits the ``Adaptation" box in Figure \ref{fig:scheme}.
	
	As introduced in \citep{Bakirov2015}, it is also possible to use $\mathcal{V}_{k}$ for the choice of $g_{{h}_k}$. Given observations, the \textit{a posteriori} prediction error $\mathcal{V}_{k}$ is $\langle(f_{k}^{-} \circ g_{{h}_k}) (\boldsymbol{X}_k),\boldsymbol{y} _k \rangle$ . However, this is effectively an in-sample error as $g_{{h}_k}$ is a function of $\{\boldsymbol{X}_k, \boldsymbol{y}_k\}$.\footnote{As a solid example consider the case where $f_{k}^{+}$ is $f_{k}^{-}$ retrained using $\{\boldsymbol{X}_k, \boldsymbol{y}_k\}$. In this case $\boldsymbol{y}_k$ are part of the training set and so we risk overfitting the model if we also evaluate the goodness of fit on $\boldsymbol{y}_k$.} To obtain a generalised estimate of the prediction error we apply q-fold\footnote{In subsequent experiments, $q=10$} cross validation. The cross-validatory adaptation strategy (denoted as \textsc{XVSelect}) uses a subset (fold), $\mathcal{S}$, of $\{\boldsymbol{X}_k, \boldsymbol{y}_k\}$ to adapt; i.e.  $f_k^{+} = f_{k}^{-} \circ g_{{h}_k}(\{\boldsymbol{X}_k, \boldsymbol{y}_k\}_{\in \mathcal{S}})$ and the remainder, $\cancel{\mathcal{S}}$, is used to evaluate, i.e. find $\langle f_{k}^{+} (\boldsymbol{X}_k)_{\in \cancel{\mathcal{S}}},\boldsymbol{y}_{k_{\in\cancel{\mathcal{S}}}}\rangle $. This is repeated $q$ times resulting in $q$ different error values and the AM, $g_{{h}_k} \in G$, with the lowest average error measure is chosen. If more than one AM has the same lowest average error, a selection among them is made randomly or utilising prior knowledge. In summary: 
	\begin{equation}
		f_{k+1}^{-}= f_k^{-} \circ g_{{h}_k}, \quad {h}_k= \argmin_{{h}_k \in 1\cdots H} \langle (f_k^{-} \circ g_{{h}_k}) (\boldsymbol{X}_k), \boldsymbol{y}_k \rangle^{\times}
	\end{equation}
	where $\langle \mbox{ } \rangle^{\times}$ denotes the cross validated error. The scheme of \textit{XVSelect} for is given in Figure \ref{fig:xvadaptscheme}.
	
	The next strategy can be used in combination with any of the above strategies as it focuses on the history of the adaptation sequence and retrospectively adapts two steps back. This is called the \textit{retrospective model correction} \citep{Bakirov2016}. Specifically, we set the current model to the output of the AM at batch $k-1$ which would have produced the best estimate in block $k$:   
	\begin{equation}
		f_{k+1}^{-}= f_{k-1}^{-} \circ g_{{h}_{k-1}}\circ g_{{h}_k}, \quad {h}_{k-1} = \argmin_{{h}_{k-1} \in 1\cdots H} \langle (f_{k-1}^{-} \circ g_{{h}_{k-1}}) (\boldsymbol{X}_k), \boldsymbol{y}_k \rangle
		\label{e:retero}
	\end{equation}
	The potential draws can be again resolved randomly or using prior knowledge\footnote{This draw resolution aspect can have a noticeable effect on the results, particularly on smaller batch sizes, where draws are more likely. For our experiments, we resolve them by having a fixed preference list of AMs, based on prior knowledge and intuition, a common situation in algorithm design. This also removes the randomness element and ensures the reproducibility of the results.}. Using the cross-validated error measure in Equation~\ref{e:retero} is not necessary, because $g_{{h}_{k-1}}$ is independent of $\boldsymbol{y}_k$. Also note the presence of $g_{{h}_k}$; retrospective correction does not in itself produce a $f_{k+1}$ and so cannot be used for prediction unless it is combined with another strategy ($g_{{h}_k}$). This strategy can be extended to consider the sequence of ${r}$ AMs while choosing the optimal state for the current batch, which we call ${r}$-step retrospective correction:
	
	\begin{multline}
		f_{k+1}^{-}= f_{k-{r}}^{-} \circ g_{{h}_{k-{r}}}\circ\cdots \circ g_{{h}_{k-1}}\circ g_{{h}_k}, \{{h}_{k-{r}}\cdots {h}_{k-1}\}= \\  = \argmin_{{h}_{k-{r}}\cdots {h}_{k-1} \in 1\cdots H} \langle (f_{k-{r}}^{-} \circ g_{{h}_{k-{r}}}\circ\cdots\circ g_{{h}_{k-1}}) (\boldsymbol{X}_k), \boldsymbol{y}_k \rangle
		\label{e:retroN}
	\end{multline}
	
	The scheme for \textit{retrospective correction} is given in Figure \ref{fig:rcadaptscheme}. Since the retrospective correction can be deployed alongside any adaptation scheme, we modify the general adaptation scheme (Figure \ref{fig:scheme}) accordingly, resulting in Figure \ref{fig:scheme2}, where Figure \ref{fig:rcadaptscheme} fits in the box ``Correction". Notice that when using this approach, the prediction function $f_k(x)$, which is used to generate predictions, can be different from the \textit{propagation} function $f_k^{'}(x)$ which is used as input for adaptation. 
	
	An important technical detail for both cross-validatory selection and retrospective correction is the resolution of draws, when two or more AMs show the same predictive performance. The draws appear frequently for classification scenarios with lower batch sizes. In these cases, a prior knowledge on AMs' predictive performance can be used to make a selection\footnote{This option is used in our experiments.}. If no such knowledge exists, a random AM, or the AM which minimises the runtime can be chosen.

	We next examine the prediction algorithms with respective adaptive mechanisms (the set $G$) used in this research.
	
	\section{Algorithms}
	\label{section:algorithm}
	
	For our experiments we have chosen the following algorithms:
	\begin{itemize}
		\item Simple  Adaptive  Batch  Local  Ensemble (SABLE) \citep{Bakirov2015},
		\item  Dynamic Weighted Majority (DWM) \citep{coltermaloof2007},
		\item Paired Learner (PL) \citep{Bach2010},
		\item Leveraged Bagging (LB) \citep{Bifet2010},
		\item BLAST \citep{Rijn2015having}.
	\end{itemize}
	
	SABLE is used to address regression problem while the other algorithms address the classification problem. We have developed batch versions of these classification algorithms, which are used in experiments. Our selection of algorithms allows to explore different types of online learning methods and different adaptive mechanisms, and demonstrate that the adaptive strategies described in this paper are in fact generic and can be applied to various adaptive algorithms with multiple AMs. Below the details of model adaptation with each algorithm are presented. 
	
	\subsection{Simple Adaptive Batch Local Ensemble (SABLE) adaptation}
	SABLE \citep{Bakirov2015} uses an ensemble of experts each  implemented using a linear model formed through Recursive Partial Least Squares (RPLS) \citep{JoeQin1998}. To get the final prediction, the predictions of base learners are combined using input/output space dependent weights (i.e. local learning), which are reflected in the \textit{descriptor} of each expert. SABLE is designed for batch streaming scenario. It  supports the creation and merger of base learners. 
	
	The SABLE algorithm allows the use of five different adaptive mechanisms (including the possibility of no adaptation). AMs are deployed as soon as the true values for the batch are available and before predicting on the next batch.  The SABLE AMs are described below\footnote{See \citep{Bakirov2017} for a full description.}. It should be noted, that as SABLE was conceived as an experimentation vehicle for AM sequences effects exploration, it does not provide a default custom adaptation strategy. 
	\begin{itemize}
		\item \textbf{SAM0} (No adaptation). No changes are applied to the predictive model, corresponding to $\emptyset$.
		
		\item \textbf{SAM1} (Batch learning).
		The simplest AM augments existing data with the data from the new batch and retrains the model. Given  predictions of each expert $f_i \in \mathcal{F}$ on $\boldsymbol{\mathcal{V}}$, $\{\hat{\boldsymbol{y}}_1,...,\hat{\boldsymbol{y}}_I\}$ and measurements of the actual values, $\boldsymbol{y}$, $\boldsymbol{\mathcal{V}}$ is partitioned into subsets in the following fashion: 
		\begin{equation}
			{z}=\underset{i\in1\cdots I}{argmin}\langle f_i({\boldsymbol{x}}_j),y_j\rangle\mbox{ }\rightarrow\mbox{ }[\boldsymbol{x}_j,y_j]\in \boldsymbol{\mathcal{V}}_{{z}}
		\end{equation}
		for every instance $[\boldsymbol{x}_j,y_j]\in \boldsymbol{\mathcal{V}}$. This creates subsets \mbox{$\boldsymbol{\mathcal{V}}_i, i=1...I$} such that $\cup_{i=1}^I\boldsymbol{\mathcal{V}}_i=\boldsymbol{\mathcal{V}}$.  Then each expert is updated using the respective dataset $\boldsymbol{\mathcal{V}}_i$.  This process updates experts only with the instances where they achieve the most accurate predictions, thus encouraging the specialisation of experts and ensuring that a single data instance is not used in the training data of multiple experts.
		
		\item \textbf{SAM2} (Batch learning with forgetting).
		This AM is similar to one above but uses decay which reduces the weight of the experts historical training data, making the most recent data more important. It is realised via RPLS update with forgetting factor $\lambda$. $\lambda$ is a hyper-parameter of SABLE.
		
		\item \textbf{SAM3} (Descriptors update / weights change).
		This AM recalculates the local descriptors using the new batch. This amounts to the change of weights of the experts.
		
		\item \textbf{SAM4} (Creation of new experts).
		New expert $s_{new}$ is created from $\boldsymbol{\mathcal{V}}_k$. Then it is checked whether the newly created expert is similar to any existing experts, in which case the older expert is removed and their descriptors are merged. Finally the descriptors of all resulting experts are updated.
		
		\item \textbf{SAM5}. SAM2 (Batch learning with forgetting) followed by SAM4 (Creation of New Experts).
	\end{itemize}
	
	\subsection{Batch Dynamic Weighted Majority (bDWM) adaptation}
	\label{bDWM}
	bDWM is an extension of DWM \citep{coltermaloof2007} designed to operate on batches of data instead of on single instances as in the original algorithm. bDWM is a global experts ensemble. Assume a set of $I$ experts $S=\{s_i,...,s_I\}$ which produce predictions $\boldsymbol{\hat{y}}=\{\hat{y}_1,...,\hat{y}_I\}$ where $\hat{y}_i=s_i(\boldsymbol{x})$ with input $\boldsymbol{x}$ and a set of all possible labels $C=\{c_1,...,c_J\}$.  Then for all $i=1\cdots I$ and $j=1\cdots J$ the matrix $A$ with following elements can be calculated: 
	\begin{equation}
		a_{i,j} =  \left\{ \begin{array}{l l}
			1 & \quad \text{if $s_{i}(\boldsymbol{x})=c_j$}\\
			0 & \quad \text{otherwise}
		\end{array}\right.
	\end{equation} 
	Assuming weights vector $\boldsymbol{w}=\{w_1,...,w_I\}$ for respective predictors in $S$, the sum of the weights of predictors which voted for label $c_j$ is $z_j=\sum\limits_{i=1}^{I}w_i a_{i,j}$.  The final prediction is\footnote{This definition is adapted from \citep{Kuncheva2004a}.}: 
	\begin{equation}
		\hat{y}=\argmax_{c_j}(z_j).
	\end{equation}

	An adaptive model based on bDWM starts with a single expert and can be adapted using an arbitrary sequence of 8 possible AMs (including no adaptation) given below.  
	\begin{itemize}
		\item \textbf{DAM0} (No adaptation). No changes are applied to the predictive model, corresponding to $\emptyset$.
		
		\item \textbf{DAM1} (Batch learning).
		After the arrival of the batch $\boldsymbol{\mathcal{V}}_t$ at time $t$ each expert is updated with it.
		
		\item \textbf{DAM2} (Weights update and experts pruning).
		Weights of experts are updated using following rule:
		\begin{equation}
			w_i^{t+1}=w_i^{t}*e^{u_i^t}.
		\end{equation}
		where $w_i^{t}$ is the weight of the $i$-th expert at time $t$, and $u_i^t$ is its accuracy on the batch $\boldsymbol{\mathcal{V}}_t$. The weights of all experts in ensemble are then normalized and the experts with a weight less than a defined threshold $\eta$ are removed. It should be noted that the choice of factor $e^{u_i^t}$ is inspired by \citet{Herbster1998}, although due to different algorithm settings, the theory developed there is not readily applicable to our scenario. Weights update is different to the original DWM, which uses an arbitrary factor $\beta<1$ to decrease the weights of misclassifying experts.
		
		\item \textbf{DAM3} (Creation of a new expert).
		New expert is created from the batch $\boldsymbol{\mathcal{V}}_t$ and is given a weight of 1.
		
		\item \textbf{DAM4}. DAM2 (Weights update and experts pruning) followed by DAM1 (Batch learning).
		
		\item \textbf{DAM5}. DAM1 (Batch learning) followed by DAM3 (Creation of a new expert).
		
		\item \textbf{DAM6}. DAM2 (Weights Update and Experts Pruning) followed by DAM3 (Creation of a new expert).
		
		\item \textbf{DAM7}. DAM2 (Weights update and experts pruning) followed by DAM1 (Batch learning) followed by DAM3 (Creation of a new expert).
	\end{itemize}
	
	\textbf{bDWM (custom adaptive strategy\footnote{To reiterate, we refer to the specific way the AMs are used in original algorithms as ``custom adaptive strategy". As the custom adaptive strategy actually defines the algorithm, we will use this term with the name of algorithm (i.e. bDWM) interchangeably.}).}
	Having presented the separate adaptive mechanisms, we now describe the bDWM, a batch version of the original DWM. It starts with a single expert with a weight of one. At time $t$, after an arrival of new batch $\boldsymbol{\mathcal{V}}_t$, experts makes predictions and overall prediction is calculated as shown earlier in this section. After the arrival of true labels all experts learn on the batch $\boldsymbol{\mathcal{V}}_t$ (invoking DAM1), update their weights (DAM2) and ensemble's accuracy $u_t$ is calculated. If $u_t$ accuracy is less than the accuracy of the naive majority classifier (based on all the batches of data seen up to this point) on the last batch, a new expert is created (DAM3). The schematics of this strategy is shown in Figure \ref{fig:bdwmbplscheme}a. This scheme fits in ``Adaptation" boxes in Figures \ref{fig:scheme} and \ref{fig:scheme2}. 
	
	\subsection{Batch Paired Learner (bPL) adaptation}
	\label{bPL}
	bPL is an extension of PL \citep{Bach2010} designed to operate on batches of data instead of on single instances as in the original algorithm. bPL maintains two learners - a \textit{stable} learner which is updated with all of incoming data and which is used to make predictions, and a \textit{reactive} learner, which is trained only on the two most recent batches. For this method, three adaptive mechanisms are available, which are described below. 
	\begin{itemize}
		\item \textbf{PAM0} (No adaptation). No changes are applied to the predictive model, corresponding to $\emptyset$.
		\item \textbf{PAM1} (Updating stable learner).
		After the arrival of the batch $\boldsymbol{\mathcal{V}}_t$ at time $t$, stable learner is updated with it.
		\item \textbf{PAM2} (Switching to reactive learner).
		Current stable learner is discarded and replaced by reactive learner.
	\end{itemize}
	
\begin{figure}
		\centering
		\begin{subfigure}[b]{0.5\linewidth}
			\centering\includegraphics[width=\textwidth]{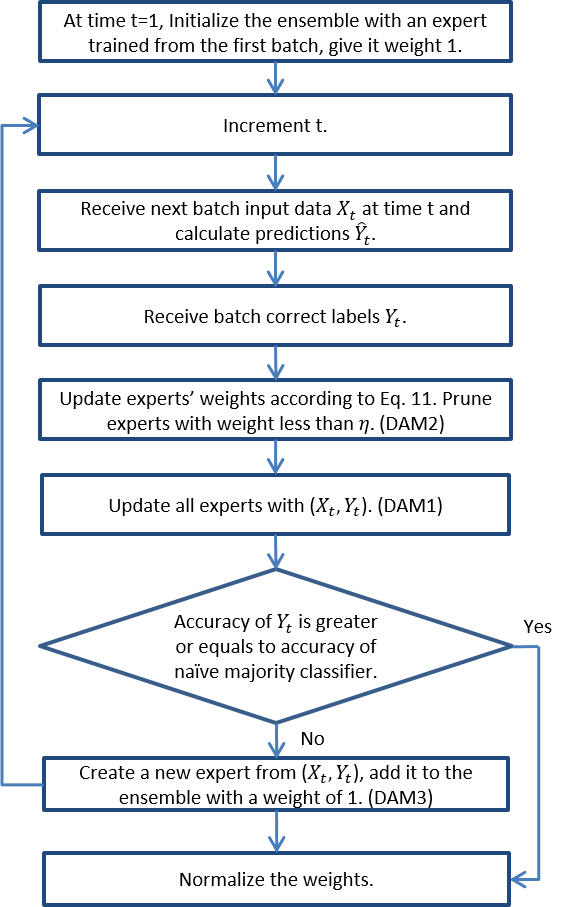}
			\caption{bDWM}
		\end{subfigure}%
		\begin{subfigure}[b]{0.5\linewidth}
			\centering\includegraphics[width=\textwidth]{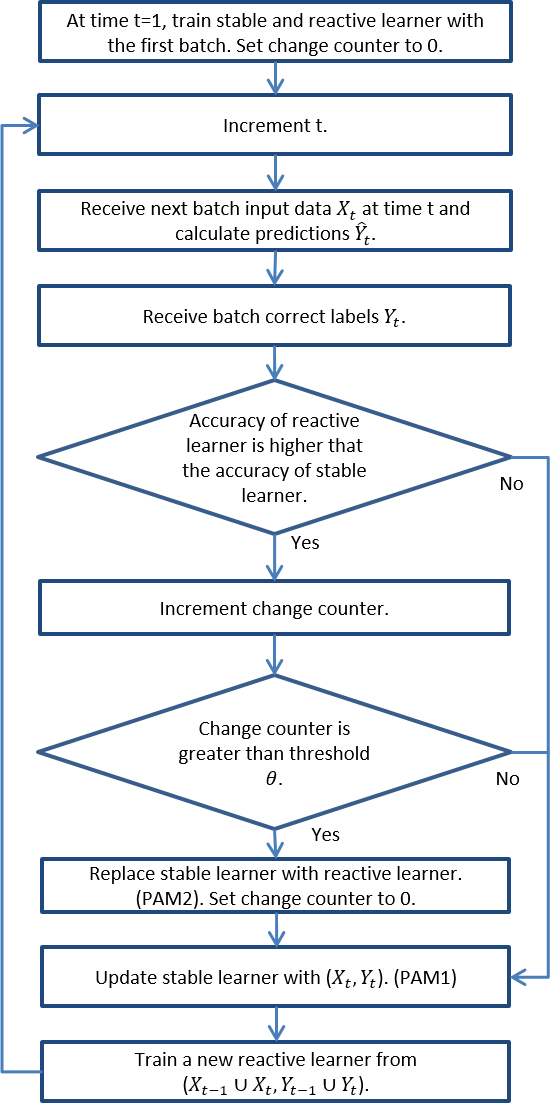}
			\caption{bPL}
		\end{subfigure}
		\caption{bDWM and bPL custom adaptation strategies.}
		\label{fig:bdwmbplscheme}
	\end{figure}

	\textbf{bPL (custom adaptive strategy).}
	Having presented the separate adaptive mechanisms, we now describe the bPL, a batch version of the original PL. Its adaptive strategy revolves around comparing the accuracy values of stable ($u_s^t$) and reactive ($u_r^t$) learners on each batch of data. Every time when $u_s^t<u_r^t$ a change counter is incremented. If the counter is higher than a defined threshold $\theta$, an existing stable learner is discarded and replaced by the reactive learner, while the counter is set to 0. As before, a new reactive learner is trained from each subsequent batch. The schematics of this strategy are shown in Figure \ref{fig:bdwmbplscheme}b. This scheme fits in ``Adaptation" boxes in Figures \ref{fig:scheme} and \ref{fig:scheme2}. 
	
	\subsection{Batch Leveraged Bagging (bLB) adaptation}
	\label{bLB}
	bLB is an extension of LB \citep{Bifet2010} designed to operate on batches of data instead of on single instances as in the original algorithm. Leveraged Bagging is based on Online Bagging \citep{Oza2001} algorithm, but includes the improvements such as the removal of experts and addition of new ones based on ADWIN \citep{Bifet2007} change detector, randomization at the ensemble output using output code etc. For this method, five adaptive mechanisms (including no change) are available, which are described below. 
	\begin{itemize}
		\item \textbf{LAM0} (No adaptation). No changes are applied to the predictive model, corresponding to $\emptyset$.
		\item \textbf{LAM1} (Batch learning).
		After the arrival of the batch $\boldsymbol{\mathcal{V}}_t$ at time $t$ each expert is updated with it.
		\item \textbf{LAM2} (Removing an existing expert and adding a new one).
		The expert with the lowest accuracy on the previously seen data is removed, and a new one trained from the most recent batch is added.
		\item \textbf{LAM3}. LAM1 (Batch learning) followed by LAM2 (Removing an existing expert and adding a new one).
	\end{itemize}
	
	\textbf{bLB (custom adaptive strategy).}
	Having presented the separate adaptive mechanisms, we now describe the bLB, a batch version of the original LB. Its strategy invokes batch learning (LAM1) after the arrival of each batch of data. If ADWIN change detector detects a change, the expert with the lowest accuracy on the previously seen data is removed, and a new one trained from the most recent batch is added (LAM2). The schematics of this strategy is shown in Figure \ref{fig:blbbblastscheme}a. This scheme fits in ``Adaptation" boxes in Figures \ref{fig:scheme} and \ref{fig:scheme2}. 
	
	\subsection{Batch BLAST (bBLAST) adaptation}
	\label{bBLAST}
	bBLAST is an extension of BLAST \citep{Rijn2015having} designed to operate on batches of data instead of on single instances as in the original algorithm. BLAST is an ensemble method using different types of base learners (as opposed to the ones mentioned above) with Online Performance Estimation for the weighting. For this method, four adaptive mechanisms (including no change) are available, which are described below. 
	\begin{itemize}
		\item \textbf{BAM0} (No adaptation). No changes are applied to the predictive model, corresponding to $\emptyset$.
		\item \textbf{BAM1} (Batch learning).
		After the arrival of the batch $\boldsymbol{\mathcal{V}}_t$ at time $t$ each expert is updated with it.
		\item \textbf{BAM2} (Reweighing the experts).
		For every instance $[\boldsymbol{x},y]\in \boldsymbol{\mathcal{V}}_t$ experts are reweighed according to Online Performance Estimation.
	\end{itemize}
	
\begin{figure}[ht]
		\centering
		\begin{subfigure}[b]{0.5\linewidth}
			\centering\includegraphics[width=\textwidth]{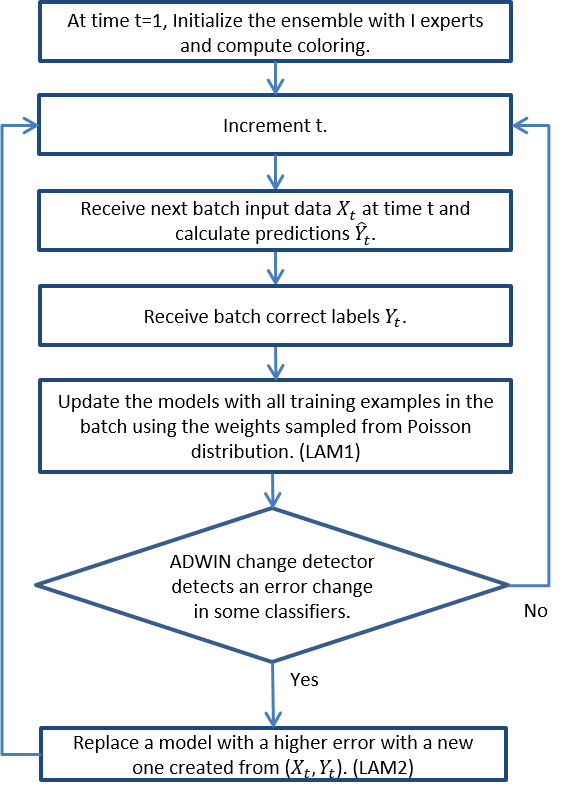}
			\caption{bLB}
		\end{subfigure}%
		\begin{subfigure}[b]{0.5\linewidth}
			\centering\includegraphics[width=\textwidth]{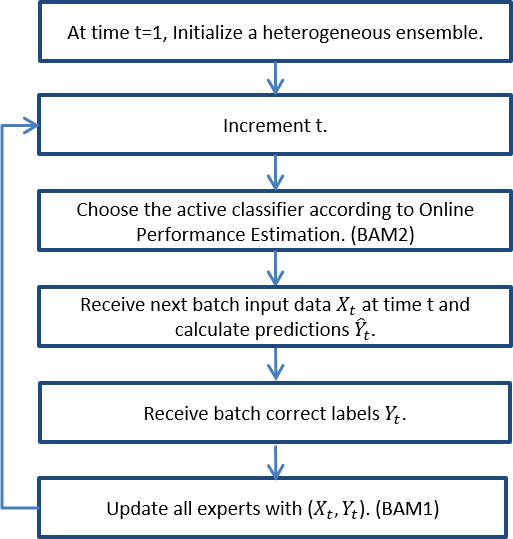}
			\caption{bBLAST}
		\end{subfigure}
		\caption{bLB and bBLAST custom adaptation strategies.}
		\label{fig:blbbblastscheme}
	\end{figure}

	\textbf{bBLAST (custom adaptive strategy).}
	Having presented the separate adaptive mechanisms, we now describe the bBLAST, a batch version of the original BLAST.
	bBLAST invokes the combination of the BAM1 (Batch learning) followed by BAM2 (Reweighing the experts) after the arrival of each batch of data. The schematics of this strategy is shown in Figure \ref{fig:blbbblastscheme}b. This scheme fits in ``Adaptation" boxes in Figures \ref{fig:scheme} and \ref{fig:scheme2}.
	
	\section{Experimental results}
	\label{section:experiments}
	In the following sub-sections we describe the empirical validation of the proposed approaches. We start by describing the experimental methodology, including experiment settings, specification of datasets, evaluation strategy, libraries and base learners used. We then follow with the comparative analysis of regression and classification results of the proposed and custom adaptive strategies.    
	
	\subsection{Methodology}
	The purpose of the experiments\footnote{All of the code except the SABLE algorithm, as well as all the datasets except Oxidizer and Drier can be found on https://github.com/RashidBakirov/multiple-adaptive-mechanisms. SABLE and the specified two datasets could not be shared because of confidentiality reasons.} in this section was to evaluate the usefulness of the proposed strategies. For this purpose we have performed the empirical comparison of automated adaptation strategies proposed in \ref{sec:strategies}  with custom adaptive strategies and with strategies involving repeated deployment of a single AM. The goal of the automated adaptive strategies is to obtain performance comparable to what one would obtain using a (usually protracted) manually optimised adaptive strategy (including hyper-parameter selection). \textit{Therefore, if the proposed strategies attain comparable, or not significantly worse accuracy levels than the custom strategies, this shall be deemed a success.}	This section discusses the results in order of introduced algorithms. For all of the algorithms we compare the MAE/accuracy of strategies listed in Table \ref{table:approaches}.
	\begin{table}
		\caption{Evaluated adaptive strategies} 
		\begin{center}
			\begin{tabularx}{\columnwidth}{rX}
				\hline
				\textbf{Result} & \textbf{Description} \\
				\hline
				\textsc{BestAM} & For all of the AMs (e.g. from DAM0 to DAM7 for the bDWM adaptation) we repeatedly deploy the same AM on all of the batches. We then select the best result among all of the runs. Note that this is a post-hoc strategy used for benchmark purposes, as the AM delivering the best result varies from dataset to dataset and is not known in advance. \\ \hline
				\textsc{BestAM+RC} & The same as \textsc{BestAM} while additionally using retrospective correction after every batch. Note that the best AM here may be different to the one from \textsc{BestAM}.\\  \hline
				\textsc{XVSelect} & Select AM (i.e. one of AMs from DAM0 to DAM7 for the bDWM adaptation) based on the current data batch using the cross-validatory approach described in Section \ref{sec:strategies}. \\ \hline
				\textsc{XVSelect+RC} & The same as \textsc{XVSelect} while additionally using retrospective correction after every batch.
				\\ \hline
				\textsc{Custom} & Using custom adaptive strategy. \\ \hline
				\textsc{Custom+RC} &  The same as \textsc{Custom} while additionally using retrospective correction after every batch.
			\end{tabularx}%
			
			\label{table:approaches}%
			
		\end{center}
	\end{table}%
	
	For SABLE, the experimentation uses five real world regression datasets listed in Table \ref{table:regressiondatareal} in Appendix \ref{appendixa}. It has been shown, e.g. in \citep{Bakirov2017,MartinSalvador2019} that these datasets present different levels of volatility and noise. For the classification algorithms, we use five real world datasets listed in Table \ref{table:classificationdatareal} and 26 synthetic datasets listed in Table \ref{table:classificationdatasynth} and visualised in Figure \ref{fig:synthdata} in Appendix \ref{appendixa}. 
	
	For the real world datasets we use prequential evaluation \citep{Dawid1984} which is a standard evaluation technique for data streams. For the batch scenario it works as follows; at time $t$ we receive the data batch $\boldsymbol{X}_t$, and predict the values/labels $\boldsymbol{\hat{y}}_t$. Then the true values/labels $\boldsymbol{y}_t$ are made available, and we calculate the error/accuracy of our predictions. Subsequently $\{\boldsymbol{X}_t, \boldsymbol{y}_t\}$ are used for adaptation. Thus, the predictions are always made on unseen data, which is not included in the training data in any form. For synthetic datasets  we generate an additional 100 test data instances for each single instance in training data using the same distribution. The predictive accuracy on the batch is then measured on test data relevant to that batch. This test data is not used for training or adapting models.
	
	For the classification algorithms, the statistical significance of differences between the results is assessed using the Friedman test with post-hoc Nemenyi test, which are widely used to compare multiple classifiers \citep{Demsar2006}. The Friedman test checks for statistical difference between the compared classifiers; if so, the Nemenyi test is used to identify which classifiers are significantly better than others. We report the results of the Nemenyi tests as Nemenyi plots\footnote{Freely available code from \citep{drawnemenyi} and \citep{Cardillo2009} were used to make these plots.}. They plot the average rank of all methods and the critical difference per batch/base learner. Classifiers that are statistically equivalent are connected by a line.
	
	For bDWM, bPL and bLB, Naive Bayes (NB)  and Hoeffding Trees (HT) \citep{Domingos2000} were used as base learners. Open source libraries Prtools \citep{PRTools}, Weka \citep{Hall2009}, MOA \citep{Bifet2010a} and scikit-multiflow \citep{Montiel2018} were employed. As there is not any randomness involved in the evaluation of datasets, a single run was used to compute the MAE (for regression) and accuracy (for classification) values, except for bLB, where 100 runs were used for each strategy.

	\subsection{Simple Adaptive Batch Local Ensemble (SABLE) results}
	
	Three different batch sizes for each dataset are examined in the simulations together using hyperparameters as tabulated in Table \ref{tab:settings}  in Appendix \ref{appendixa}. These parameter combinations were empirically identified using grid search, optimising the performance of the \textsc{Oracle} strategy (Eq. \ref{eq:oracle}). 
	
	The results of the experiments using SABLE for batch sizes $n=50, 100, 200$ are given in Table \ref{tab:sableresults}. These results suggest that most of the times \textsc{XVSelect} and \textsc{XVSelect+RC} perform better or comparable to \textsc{BestAM} and \textsc{BestAM+RC}. Overall \textsc{XVSelect} or \textsc{XVSelect+RC} had the lowest MAE with significant difference in 7 experiments out of 15. \textsc{XVSelect} or \textsc{XVSelect+RC} showed comparable (not worse with significant difference) performance to \textsc{BestAM} in 11 experiments. The cases where \textsc{XVSelect} and \textsc{XVSelect+RC} perform noticeably worse are Drier dataset with batch size of 100 and Sulfur dataset with all batch sizes. We relate this to the stability of these datasets. Indeed, the \textsc{BestAM} in all these cases is the slow adapting sequence of SAM1, without any forgetting of the old information. Difference in batch sizes is important for some datasets. This can be related to the frequency of changes and whether they happen within a batch, which can have a negative impact on \textsc{XVSelect} and \textsc{XVSelect+RC}. Retrospective correction (RC) has improved the performance of \textsc{XVSelect} for some cases. For the deployment of single AM, as seen in \textsc{BestAM} and \textsc{BestAM+RC} results, RC is more useful for the larger batch sizes, presumably because more training data prevents overfitting.
	
	\begin{table}[htbp]
		\centering
		\captionsetup{singlelinecheck=false}
		\caption{SABLE results. The best performance in each row is indicated with bold font. The AM which was found to deliver the best for  performance for \textsc{BestAM} and \textsc{BestAM+RC} is indicated in respective columns. Upwards arrow denotes the cases when either \textsc{XVSelect} or \textsc{XVSelect+RC} performs better that \textsc{BestAM}, and downwards arrow denotes the opposite cases. Double lined arrows indicate a significant difference according to Wilcoxon signed-rank test\tablefootnote{The Wilcoxon signed-rank test assumes the null distribution is symmetric. This assumption mostly holds for our data.} \citep{Wilcoxon1945Individual} with $p=0.05$.}
		\begin{tabular}{lrrrr}
			&&\multicolumn{1}{c}{$n=50$}&&\\ \hline
			& \multicolumn{1}{l}{\textsc{BestAM}} & \multicolumn{1}{l}{\textsc{BestAM+RC}} & \multicolumn{1}{l}{\textsc{XVSelect}} & \multicolumn{1}{l}{\textsc{XVSelect+RC}} \\ \hline
			\textbf{Catalyst}  	$\Uparrow$ & 0.023 (SAM2) & 0.028 (SAM5) & \textbf{0.021} & 0.023 \\
			\textbf{Oxidiser $\Uparrow$} & 0.490 (SAM2) & 0.501 (SAM4) & \textbf{0.485} & 0.519 \\
			\textbf{Drier} $\Uparrow$& 8.98$\times$10$^{-6}$ (SAM1) & 9.78$\times$10$^{-6}$ (SAM0) & 9.27$\times$10$^{-6}$ & \textbf{6.95$\times$10$^{-6}$} \\
			\textbf{Debutaniser} $\downarrow$& \textbf{0.117} (SAM1)& 0.121 (SAM4) & 0.122 & 0.122 \\
			\textbf{Sulfur $\Downarrow$} & \textbf{0.030} (SAM1) & 0.051 (SAM3) & 0.060 & 0.050 \\ \hline
			\\
			&&\multicolumn{1}{c}{$n=100$}&&\\ \hline
			& \multicolumn{1}{l}{\textsc{BestAM}} & \multicolumn{1}{l}{\textsc{BestAM+RC}} & \multicolumn{1}{l}{\textsc{XVSelect}} & \multicolumn{1}{l}{\textsc{XVSelect+RC}} \\ \hline
			\textbf{Catalyst} $\Uparrow$ & 0.031 (SAM2) & 0.031 (SAM4) & 0.030 & \textbf{0.029} \\
			\textbf{Oxidiser} $\Downarrow$& \textbf{0.542} (SAM4) & 0.559 (SAM4) & 0.569 & 0.566 \\
			\textbf{Drier} $\Downarrow$& \textbf{8.09$\times$10$^{-6}$} (SAM1) & 8.97$\times$10$^{-6}$ (SAM1) & 1.20$\times$10$^{-5}$ & 1.12$\times$10$^{-5}$ \\
			\textbf{Debutaniser} $\Uparrow$ & 0.117 (SAM1) & 0.116 (SAM4) & 0.145 & \textbf{0.112} \\
			\textbf{Sulfur} $\Downarrow$& \textbf{0.031} (SAM1) & 0.058 (SAM2) & 0.060 & 0.054 \\ \hline
			\\
			&&\multicolumn{1}{c}{$n=200$}&&\\ \hline
			& \multicolumn{1}{l}{\textsc{BestAM}} & \multicolumn{1}{l}{\textsc{BestAM+RC}} & \multicolumn{1}{l}{\textsc{XVSelect}} & \multicolumn{1}{l}{\textsc{XVSelect+RC}} \\ \hline
			\textbf{Catalyst} $\Uparrow$ & 0.0495 (SAM4) & 0.0519 (SAM5) & \textbf{0.0492} & 0.0495 \\
			\textbf{Oxidiser} $\downarrow$ & 0.612 (SAM4) & \textbf{0.611} (SAM5) & 0.631 & 0.676 \\
			\textbf{Drier} $\Uparrow$ & 5.01$\times$10$^{-5}$ (SAM4) & 5.01$\times$10$^{-5}$ (SAM5) & \textbf{4.67$\times$10$^{-5}$} & \textbf{4.67$\times$10$^{-5}$} \\
			\textbf{Debutaniser} $\uparrow$& 0.106 (SAM1) & 0.105 (SAM4) & \textbf{0.104} & 0.108 \\
			\textbf{Sulfur} $\Downarrow$ & \textbf{0.033} (SAM1) & 0.039 (SAM1) & 0.049 & 0.040 \\
			\hline
		\end{tabular}%
		\label{tab:sableresults}%
	\end{table}%
	
	\subsection{Batch Dynamic Weighted Majority (bDWM) results}
	\label{sec:bdwm_results}
\begin{figure}[ht]
		\centering
		\begin{subfigure}[b]{0.5\linewidth}
			\centering\includegraphics[width=\textwidth]{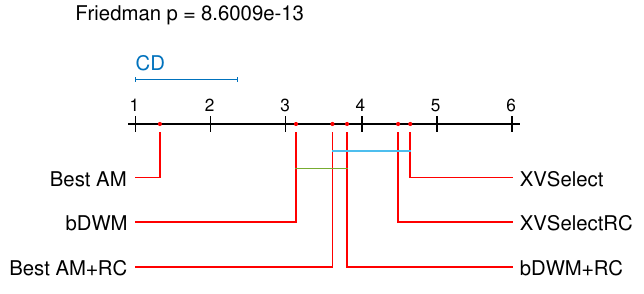}
			\caption{Base learner: NB, $n=10$}
		\end{subfigure}%
		\begin{subfigure}[b]{0.5\linewidth}
			\centering\includegraphics[width=\textwidth]{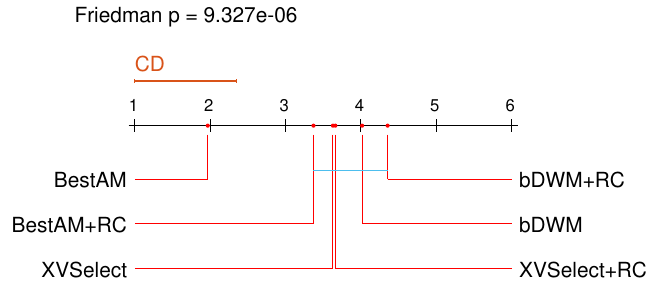}
			\caption{Base learner: HT, $n=10$}
		\end{subfigure}%
		
		\begin{subfigure}[b]{0.5\linewidth}
			\centering\includegraphics[width=\textwidth]{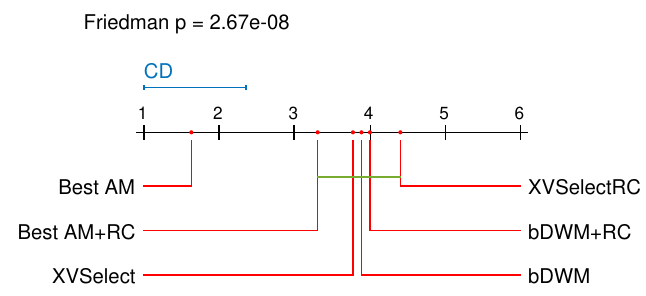}
			\caption{Base learner: NB, $n=20$}
		\end{subfigure}%
		\begin{subfigure}[b]{0.5\linewidth}
			\centering\includegraphics[width=\textwidth]{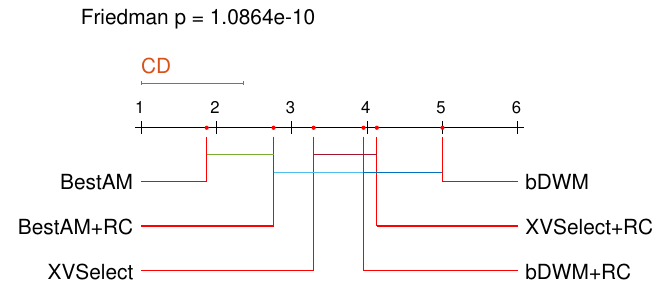}
			\caption{Base learner: HT, $n=20$}
		\end{subfigure}%
		
		\begin{subfigure}[b]{0.5\linewidth}
			\centering\includegraphics[width=\textwidth]{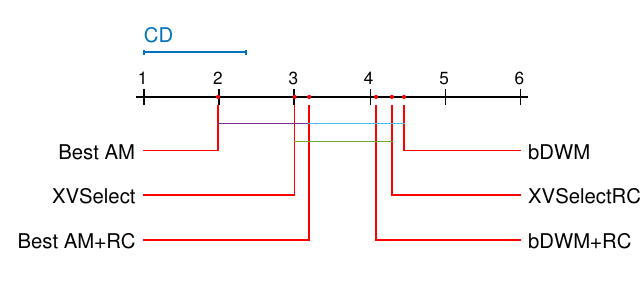}
			\caption{Base learner: NB, $n=50$}
		\end{subfigure}%
		\begin{subfigure}[b]{0.5\linewidth}
			\centering\includegraphics[width=\textwidth]{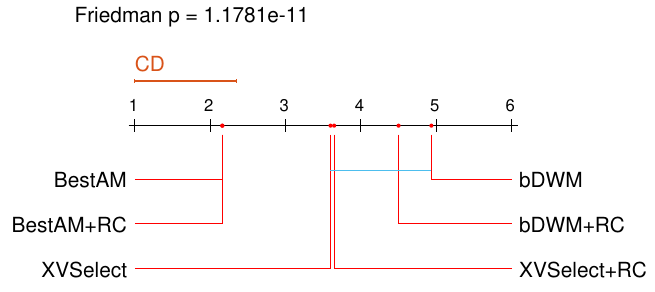}
			\caption{Base learner: HT, $n=50$}
		\end{subfigure}%
		\caption{bDWM adaptation: Nemenyi plots (lower is better) of \textsc{BestAM}, \textsc{BestAM+RC}, \textsc{XVSelect}, \mbox{\textsc{XVSelect+RC}}, \textsc{Custom} (bDWM), \textsc{Custom+RC} (bDWM+RC) strategies for different batch sizes $n$ with NB and HT base learners.}
		\label{fig:bdwm_nemenyi}
	\end{figure}

	The results of the Nemenyi test are shown
	in Figure \ref{fig:bdwm_nemenyi}\footnote{The full results tables with accuracy values of each approach on each dataset are accessible from https://github.com/RashidBakirov/multiple-adaptive-mechanisms/tree/master/results.}. For four experiments out of six, excluding NB base learner with batch sizes of 10 and 20, \textsc{XVSelect} and \textsc{XVSelect+RC} are both ranked higher than the bDWM (\textsc{Custom} strategy), in some cases significantly so. For batch size 10 with NB as base learner, bDWM performs better than both proposed approaches and for batch size 20, better than \textsc{XVSelect+RC}. The addition of retrospective correction does not seem to bring obvious benefit to adaptive strategies; while improving the results in some experiments, in most of the cases it decreases the accuracy. In terms of batch sizes, increasing $n$ seems to improve the performance of \textsc{XVSelect} with NB base learner. In general, \textsc{BestAM} provides the best results across all experiments, while \textsc{BestAM+RC} performs slightly worse. It may be worth to reiterate that, for all of the classification experiments, the \textsc{BestAM} and \textsc{BestAM+RC} repeatedly deploy the single AM which delivers the best results \textit{specific for particular settings (dataset, batch size, base learner)}. This AM is not known in advance, so this strategy is not attainable in practice and is used for benchmark purposes.  
	
	\subsection{Batch Paired Learner (bPL) results}
	
\begin{figure}[ht]
		\centering
		\begin{subfigure}[b]{0.5\linewidth}
			\centering\includegraphics[width=\textwidth]{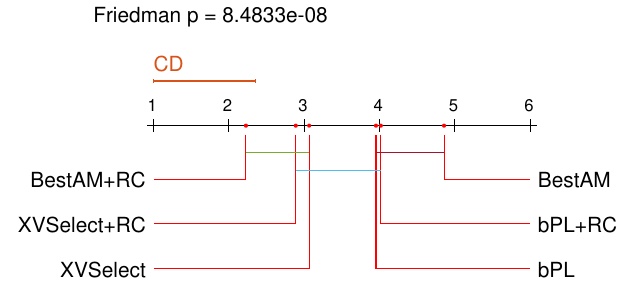}
			\caption{Base learner: NB, $n=10$}
		\end{subfigure}%
		\begin{subfigure}[b]{0.5\linewidth}
			\centering\includegraphics[width=\textwidth]{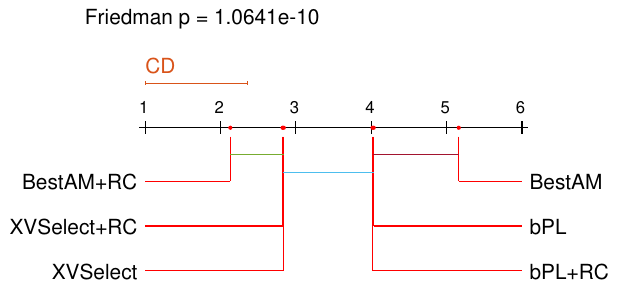}
			\caption{Base learner: HT, $n=10$}
		\end{subfigure}%
		
		\begin{subfigure}[b]{0.5\linewidth}
			\centering\includegraphics[width=\textwidth]{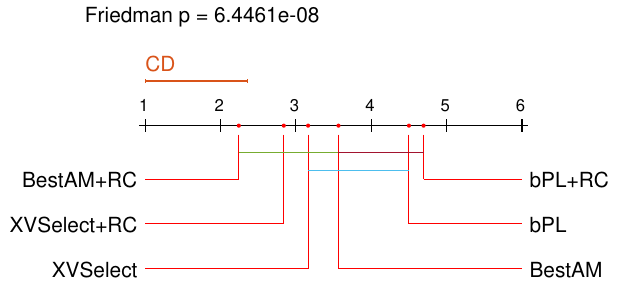}
			\caption{Base learner: NB, $n=20$}
		\end{subfigure}%
		\begin{subfigure}[b]{0.5\linewidth}
			\centering\includegraphics[width=\textwidth]{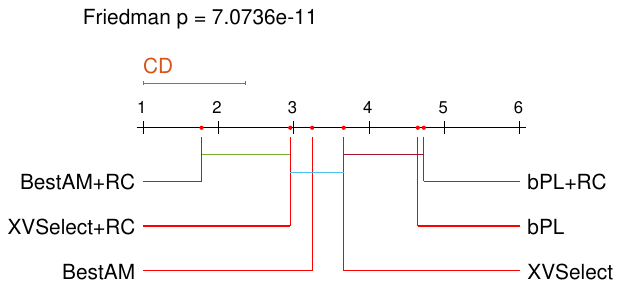}
			\caption{Base learner: HT, $n=20$}
		\end{subfigure}%
		
		\begin{subfigure}[b]{0.5\linewidth}
			\centering\includegraphics[width=\textwidth]{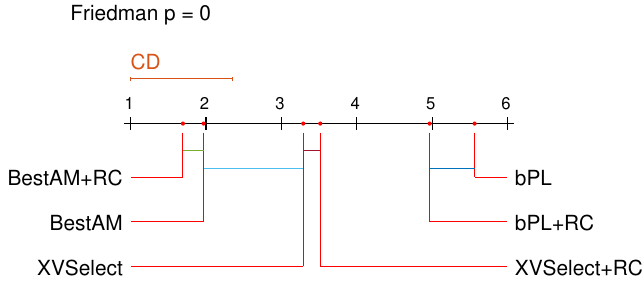}
			\caption{Base learner: NB, $n=50$}
		\end{subfigure}%
		\begin{subfigure}[b]{0.5\linewidth}
			\centering\includegraphics[width=\textwidth]{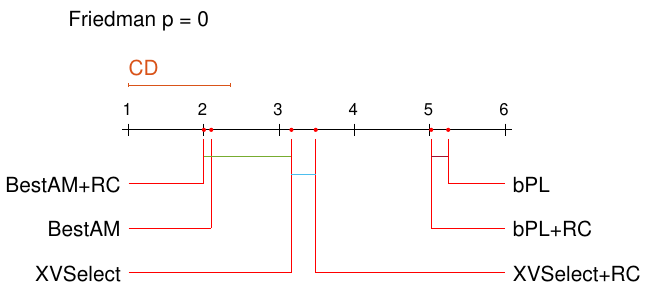}
			\caption{Base learner: HT, $n=50$}
		\end{subfigure}%
		\caption{bPL adaptation: Nemenyi plots (lower is better) of \textsc{BestAM}, \textsc{BestAM+RC}, \textsc{XVSelect}, \mbox{\textsc{XVSelect+RC}}, \textsc{Custom} (bPL), \textsc{Custom+RC} (bPL+RC) strategies for different batch sizes $n$ with NB and HT base learners.}
		\label{fig:bpl_nemenyi}
	\end{figure}

	For bPL and bPL+RC (\textsc{Custom} and \textsc{Custom+RC} strategies) we have used the threshold of $\theta=1$ for all the experiments. This value was chosen as it was experimentally established that the lower threshold values tend to provide better results than the higher ones. At the same time, keeping $\theta>0$ makes use of the change counter mechanism, a characteristic feature of bPL ($\theta=0$ provided similar results). We present the Nemenyi plots for both base learners on all three batch sizes in Figure \ref{fig:bpl_nemenyi}. Also for this algorithm, \textsc{XVSelect} and \textsc{XVSelect+RC} show good performance and are ranked higher than the bPL for all batch sizes and base learner combinations. For bPL adaptation, the \textsc{BestAM+RC} performs well for all of the settings, however the performance of \textsc{BestAM} is poor for the low batch sizes. Retrospective correction appears to be useful for bPL adaptation, providing improvements for \textsc{BestAM} and \textsc{XVSelect} for most settings.
	
	\subsection{Batch Leveraged Bagging (bLB) results}
\begin{figure}[ht]
		\centering
		\begin{subfigure}[b]{0.5\linewidth}
			\centering\includegraphics[width=\textwidth]{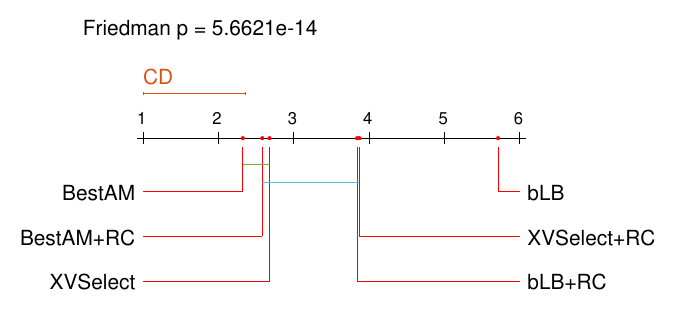}
			\caption{Base learner: NB, $n=10$}
		\end{subfigure}%
		\begin{subfigure}[b]{0.5\linewidth}
			\centering\includegraphics[width=\textwidth]{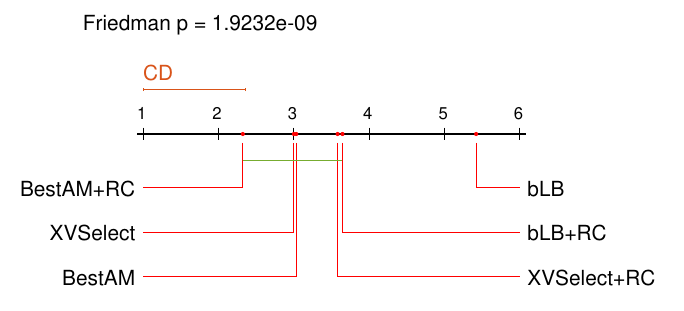}
			\caption{Base learner: HT, $n=10$}
		\end{subfigure}%
		
		\begin{subfigure}[b]{0.5\linewidth}
			\centering\includegraphics[width=\textwidth]{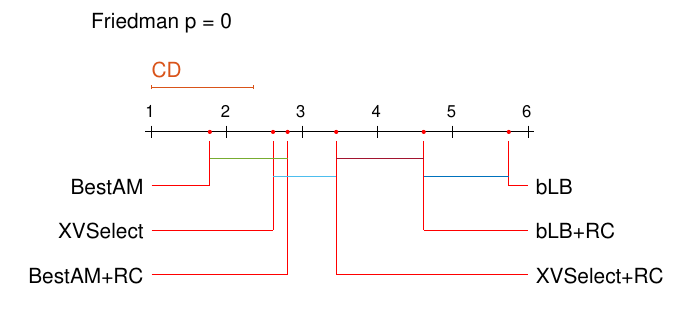}
			\caption{Base learner: NB, $n=20$}
		\end{subfigure}%
		\begin{subfigure}[b]{0.5\linewidth}
			\centering\includegraphics[width=\textwidth]{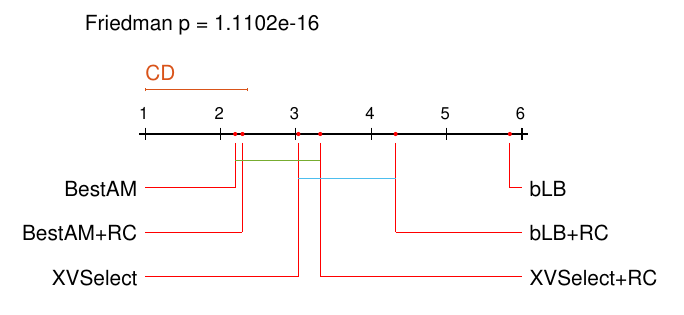}
			\caption{Base learner: HT, $n=20$}
		\end{subfigure}%
		
		\begin{subfigure}[b]{0.5\linewidth}
			\centering\includegraphics[width=\textwidth]{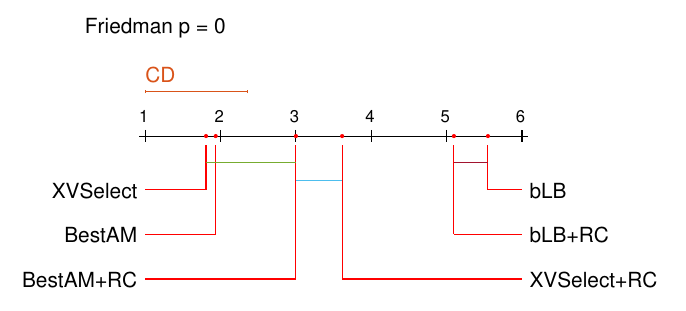}
			\caption{Base learner: NB, $n=50$}
		\end{subfigure}%
		\begin{subfigure}[b]{0.5\linewidth}
			\centering\includegraphics[width=\textwidth]{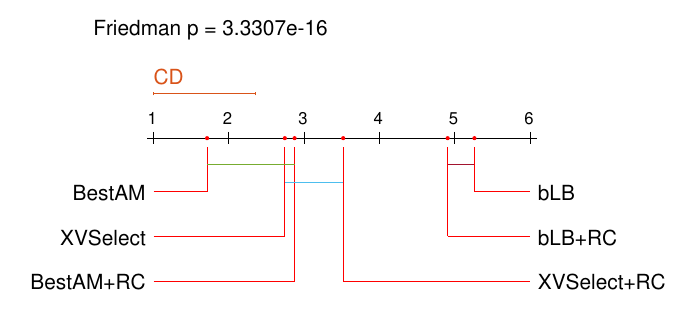}
			\caption{Base learner: HT, $n=50$}
		\end{subfigure}%
		\caption{bLB adaptation: Nemenyi plots (lower is better) of  \textsc{BestAM}, \textsc{BestAM+RC}, \textsc{XVSelect}, \mbox{\textsc{XVSelect+RC}}, \textsc{Custom} (bLB), \textsc{Custom+RC} (bLB+RC) strategies for different batch sizes $n$ with NB and HT base learners.}
		\label{fig:blb_nemenyi}
	\end{figure}

	bLB adaptation was implemented modifying the existing code from scikit-multiflow. The default hyper-parameters were kept. We present the Nemenyi plots of the average accuracy values of 100 runs for each adaptive strategy for both base learners on all three batch sizes in Figure \ref{fig:blb_nemenyi}. The performance of the proposed \textsc{XVSelect} is consistently better than the bLB (\textsc{Custom} strategy) for all of the settings, mostly significantly so. This is even more apparent for higher batch sizes. Behaviour of RC in this case is noteworthy; \textsc{XVSelect+RC} performs consistently worse than \textsc{XVSelect} although still beats the bLB in all of the settings bar one. On the other hand, bLB with RC (\textsc{Custom+RC} strategy) is always better than the bLB. It is possible that for Leveraged Bagging, combining \textsc{XVSelect} and RC makes the adaptation overfit to the last batch, thus reducing the accuracy. For bLB adaptation, the \textsc{BestAM} outperforms the proposed approaches in most of the settings, however there are no significant differences to the performance of \textsc{XVSelect}. 
	
	\begin{figure}[ht]
		\centering
		\begin{subfigure}[b]{0.5\linewidth}
			\centering\includegraphics[width=\textwidth]{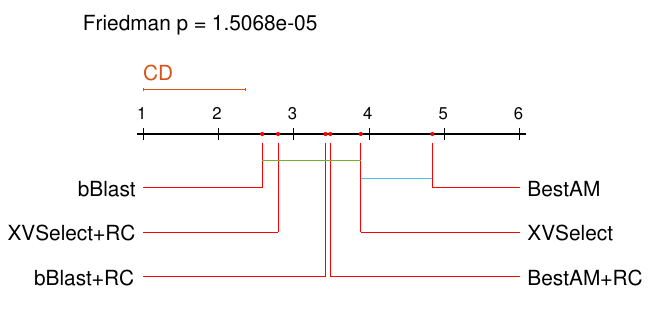}
			\caption{$n=10$}
		\end{subfigure}%
		\begin{subfigure}[b]{0.5\linewidth}
			\centering\includegraphics[width=\textwidth]{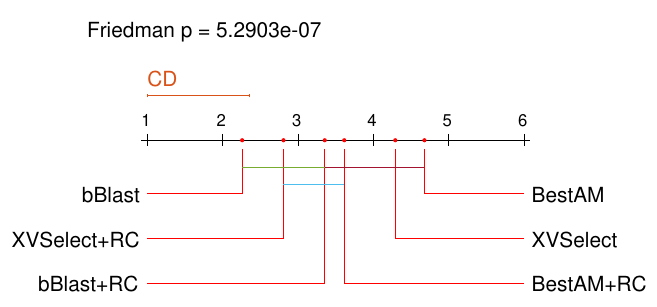}
			\caption{$n=20$}
		\end{subfigure}
		\begin{subfigure}[b]{0.5\linewidth}
			\centering\includegraphics[width=\textwidth]{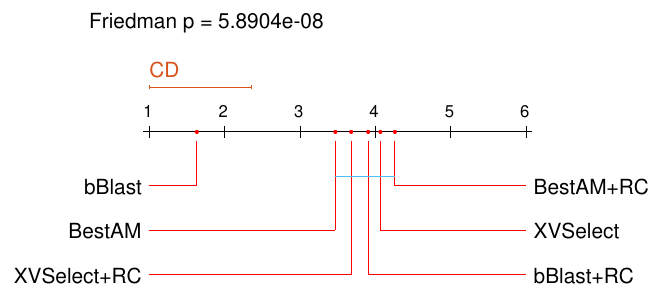}
			\caption{$n=50$}
		\end{subfigure}
		\caption{bBLAST adaptation: Nemenyi plots (lower is better) of \textsc{BestAM}, \textsc{BestAM+RC}, \textsc{XVSelect}, \mbox{\textsc{XVSelect+RC}}, \textsc{Custom} (bBLAST), \textsc{Custom+RC} (bBLAST+RC) adaptive strategies for different batch sizes $n$.}
		\label{fig:blast_nemenyi}
	\end{figure}
	
	\subsection{Batch BLAST (bBLAST) results}
	\label{sec:bblast_results}
	bBLAST adaptation was implemented modifying the existing MOA code. In contrast to the algorithms in the previous sections, bBLAST uses not single but multiple base learning algorithms; Hoeffding Tree, Naive Bayes, Perceptron, Stochastic Gradient Descent, and k Nearest Neighbour. All of the parameters of the bBLAST, as well as those of base experts are kept at defaults of MOA.
	We present the Nemenyi plots of the average the accuracy values of the selected adaptive strategies for all three batch sizes in Figure \ref{fig:blast_nemenyi}. The  performance of the bBLAST (\textsc{Custom} strategy) is consistently better than the proposed adaptive strategies for all of the settings, though not significantly different than \textsc{XVSelect+RC} for batch sizes $n=10$ and $n=20$. The RC effect here is the mirror opposite to the bLB; bBLAST with RC (\textsc{Custom+RC}) always performs worse than bBLAST, however \textsc{XVSelect+RC} always performs better than \textsc{XVSelect}. Performance of \textsc{BestAM} and \textsc{BestAM+RC} strategies for this algorithm is markedly worse than for others as they are often outperformed by \textsc{XVSelect} and \textsc{XVSelect+RC}.

	\subsection{Summary of classification results}
	The conducted experiments give insight on several questions. Firstly, we are interested whether the proposed adaptation strategies \textsc{XVSelect} and \textsc{XVSelect+RC} provide comparable results to the custom strategies or to the best results achieved by a repeated deployment of any AM. Secondly, we would like to know whether the retrospective correction has really a positive effect on the accuracy of the predictions, and if so for which approaches. Finally, we would like to compare the performance of the adaptive strategies on the synthetic data to this on real-world datasets. To answer the first two questions we compare the results from sections \ref{sec:bdwm_results}-\ref{sec:bblast_results} in Table \ref{table:comparison}, summing up the number of cases one approach was better and worse than the other across all of the algorithms, batch sizes and base learners (equal performance is represented by 0.5 in both ``Better" and ``Worse" columns). 
	
	\begin{table}[htbp]
		\centering
		\caption{Comparisons of different approaches}
		\begin{tabular}{lcccc}
			\hline
			Comparison & \multicolumn{1}{l}{Better (significant)} & \multicolumn{1}{l}{Better} & \multicolumn{1}{l}{Worse} & \multicolumn{1}{l}{Worse (significant)} \\ \hline
			\textsc{XVSelect} vs \textsc{Custom} & 11     & 6     & 1     & 3 \\
			\textsc{XVSelect+RC} vs \textsc{Custom} & 10     & 6     & 3     & 2 \\
			\textsc{XVSelect} vs \textsc{BestAM} & 2     & 5     & 9     & 5 \\
			\textsc{XVSelect+RC} vs \textsc{BestAM} & 4     & 2     & 3     & 12 \\
			\textsc{XVSelect+RC} vs \textsc{XVSelect} & 1     & 8    & 11     & 1 \\
			\textsc{Custom+RC} vs \textsc{Custom} & 3     & 9    & 8     & 1 \\
			\textsc{BestAM+RC} vs \textsc{BestAM}   & 4     & 5.5     & 8.5     & 3 \\ \hline
		\end{tabular}%
		\label{table:comparison}%
	\end{table}%

	In comparison to \textsc{Custom}, \textsc{XVSelect} and \textsc{XVSelect+RC} has better accuracy for most experiments, often with significant difference. For these comparisons \textsc{XVSelect} and \textsc{XVSelect+RC} show similar results. Both \textsc{XVSelect} or \textsc{XVSelect+RC} perform in average worse than \textsc{BestAM}, however, for \textsc{XVSelect}, the performance is comparable (not significantly worse in majority of cases).
	
	Furthermore, we consider the effects of RC separately for each approach, as it has been shown that they could be different. For \textsc{XVSelect}, deploying RC seems to not have a critical effect. The positive effect of RC is more apparent on \textsc{Custom} strategy. For the \textsc{BestAM} it should be noted that the best AM can be different depending on dataset and even for the same dataset it is not necessarily the case that \textsc{BestAM} and \textsc{BestAM+RC} will be based on the same AM. However, we can still say if the best performing AM is known, the deployment of RC is likely to have a negative effect on the accuracy.
	
\begin{figure}
		\centering
		\begin{subfigure}[b]{0.5\linewidth}
			\centering\includegraphics[width=\textwidth]{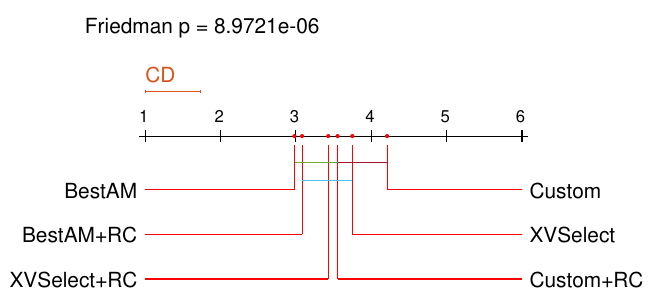}
			\caption{Real data}
		\end{subfigure}%
		\begin{subfigure}[b]{0.5\linewidth}
			\centering\includegraphics[width=\textwidth]{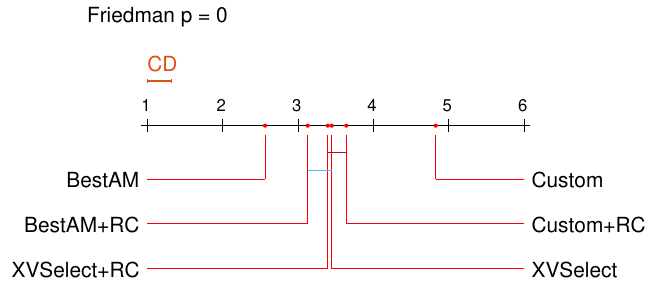}
			\caption{Synthetic data}
		\end{subfigure}
		\caption{Nemenyi plots (lower is better) of \textsc{BestAM}, \textsc{BestAM+RC}, \textsc{XVSelect}, \mbox{\textsc{XVSelect+RC}}, \textsc{Custom}, \textsc{Custom+RC}  strategies for real and synthetic datasets.}
		\label{fig:classnemenyi}
	\end{figure}

	Finally, to evaluate the performance of \textsc{XVSelect} and \textsc{XVSelect+RC} on the synthetic vs. real world data, we have compared the results on these datasets separately, across all of the algorithms and settings using Nemenyi plots on Figure \ref{fig:classnemenyi}. It is possible to observe that the results of the proposed approaches is closer to the \textsc{BestAM} on the real world data, with \textsc{XVSelect}, \textsc{XVSelect+RC} and \textsc{Custom+RC} showing comparable performance. This may be related to the more complicated nature of these datasets, where there may not exist a single AM that markedly optimises the performance, an observation in line with our earlier findings from \citep{Bakirov2015}. The performance of \textsc{XVSelect} and \textsc{XVSelect+RC} is comparatively worse on synthetic data, which may be simple enough for a single AM based adaptive strategy to deliver good results. Even for this case, these two approaches outperform \textsc{Custom} with a significant difference.
	
	\subsection{Runtime analysis}
	
	We proceed with the analysis of the runtime performance of our approaches. First, we note that with the assumption that the processing time for every batch, including the prediction, adaptation, and accuracy/error calculation is bounded by some constant, which is the case for all of the algorithms we consider, the runtime complexity of any custom adaptive algorithm is $\mathcal{O}(n)$, where $n$ is the number of batches. In this case, the runtime complexity of \textsc{XVSelect} is $\mathcal{O}(|G|qn)$ where $|G|$ is the number of available AMs and $q$ is the number of cross-validation fold, as for every batch every AM with $q$-fold cross-validation is used. Retrospective correction has the complexity of $\mathcal{O}(|G|n)$, as for every batch every AM is used once. Thus, \textsc{XVSelect+RC} has the complexity of $\mathcal{O}(|G|^{2}qn)$. Since $|G|$ and $q$ are constants, it follows that $\mathcal{O}(|G|n)\sim\mathcal{O}(|G|qn)\sim\mathcal{O}(|G|^{2}qn)\sim\mathcal{O}(n)$, hence all the proposed methods are in the same order of runtime complexity as the custom strategies.  
	
	For empirical runtime evaluation, we compare the performance of \textsc{XVSelect}, \mbox{\textsc{XVSelect+RC}}, \textsc{Custom}, \textsc{Custom+RC} strategies on classification dataset \#28 (Power Italy) for different classification algorithms with $n=50$ and NB base learner in Table \ref{tab:runtimes}\footnote{The results in this section are achieved on quad-core Intel Core i7-7700HQ CPU with core frequency of 2.8 GHz. All adaptive strategies were run 10 times and the average results are reported.}, initially without using any parallel processing. This dataset was chosen as it is a relatively large sized real-world dataset.  The results show that the performance of our methods vary greatly depending on algorithm; e.g. for \textsc{XVSelect+RC} with 2-fold cross-validation, bPL adaptation has fastest relative average batch processing time (only 2.69 times higher than \textsc{Custom}), whereas bDWM adaptation has the slowest time (110.73 times higher than \textsc{Custom}).
	
	The differences in performances are explainable by the internal characteristics of the algorithms. Batch processing time for \textsc{XVSelect} and \mbox{\textsc{XVSelect+RC}} is proportional to the batch runtimes with single AMs (e.g. when using \textsc{Custom} strategy). The longer batch runtimes are further extended by the cross-validation and retrospective correction. Therefore, \textsc{XVSelect} and \textsc{XVSelect+RC} for bDWM which has 8 AMs and can have about 20 active experts at the same time, has much higher relative batch runtime than bPL, which has only three AMs and two experts. Other interesting observation is that the RC does not always increase the batch processing time as seen in the example of bPL, which inherently deploys all of the AMs even without RC. This is also the case for bDWM \textsc{XVSelect} and \mbox{\textsc{XVSelect+RC}}, where this may be attributed to the AMs deployed by \mbox{\textsc{XVSelect+RC}} strategy (e.g. less creation of new experts AMs, which notably slow the model down).
	
	Batch processing runtime can be improved by applying parallel processing as both cross-validatory selection and retrospective correction are embarrassingly parallel operations. Fully parallelising the adaptive strategy however requires available $|G|q$ threads which can be prohibitive. Even the fully parallel implementation may not be as efficient as the custom strategy, because the choice of the AM can have an effect on the performance for the subsequent batches. This can be again seen on an example of expert creation AMs. 
	
\begin{figure}[!h]
		\centering\includegraphics[width=0.6\textwidth]{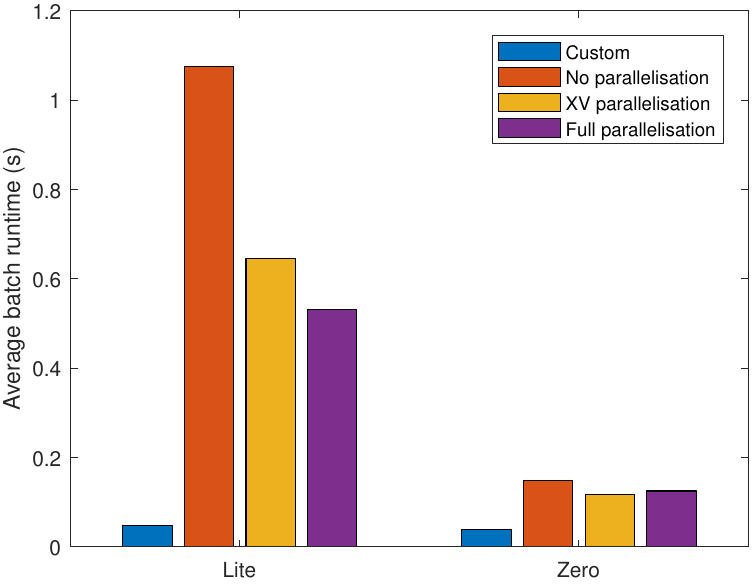}
		\caption{Average batch runtimes for bDWM Lite and Zero, \textsc{XVSelect+RC} strategy on classification dataset \#28 (Power Italy) with $n=50$ and NB base learner.}
		\label{fig:runtimesbar}%
	\end{figure}

\begin{figure}[!h]
		\centering\includegraphics[width=\textwidth]{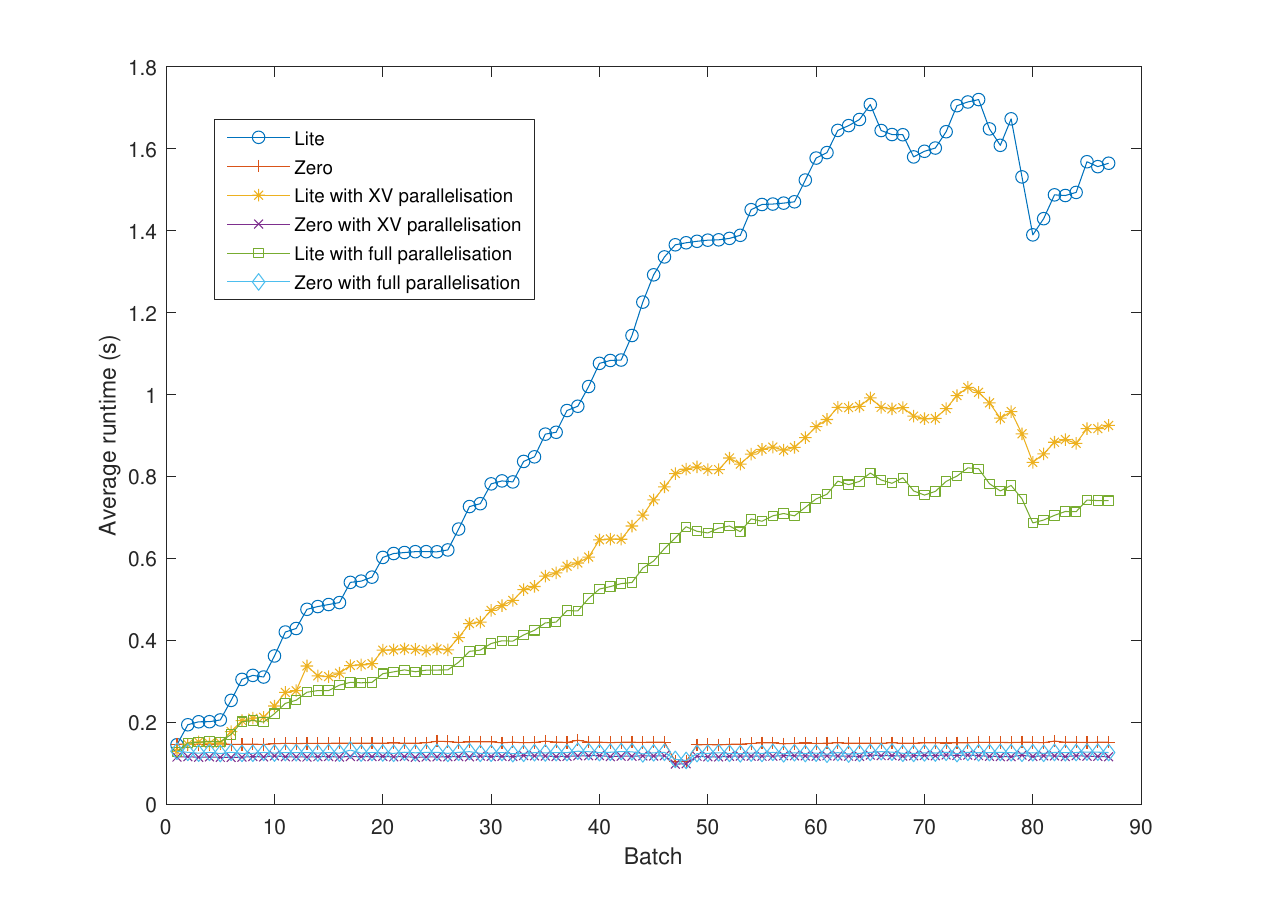}
		\caption{Average batch runtimes for bDWM Lite and Zero, \textsc{XVSelect+RC} strategy on classification dataset \#28 (Power Italy) with $n=50$ and NB base learner.}
		\label{fig:runtimesplot}%
	\end{figure}

	To illustrate these points a further experiment is undertaken, where two modifications of bDWM are proposed. The first one, bDWM\_Lite starts with two experts and includes only two AMs, DAM4 (weights  update, experts  pruning and batch learning) and DAM7 (weights  update, experts pruning, batch learning and expert creation) instead of the original 8, which still allows to run the \textsc{Custom} strategy. bDWM\_Lite allows us to test the fully parallel implementation as it requires only 4 threads for this. The second modification, bDWM\_Zero, mimics bDWM\_Lite, and in addition limits the ensemble to only two experts. This prevents the performance degradation caused by expert creation. We experiment with \textsc{XVSelect+RC} with 2-fold cross-validation and two parallelisation\footnote{Parallelisation is realised using Matlab Parallel toolbox.} choices, cross-validation (XV) parallelisation where parallel processing is applied to the cross-validation only, and full parallelisation, where in addition to cross-validation, the retrospective correction is also run in parallel. Figure \ref{fig:runtimesbar} shows the average batch runtimes over the whole dataset. Even without parallelisation, simply reducing the number of AMs from 8 to 2 (bDWM\_Lite), results in performance increase by the factor of 6, while parallelisation increases it even further. Limiting the number of experts further reduces the average batch runtime to only 3 times more than the \textsc{Custom}. Note that for bDWM\_Zero the parallelisation does not decrease the runtimes by much and that the full parallelisation doesn't outperform XV only parallelisation. This can be attributed to the already reduced runtime due to limited number of experts and the parallel processing overhead which negates increase in performance. Further insights are given in Figure \ref{fig:runtimesplot}. It can be seen that for bDWM\_Lite, average runtime per batch increases as batches come in, due to the increase in  experts, however gradually flattens as the number of experts stabilizes around 20. Conversely, for bDWM\_Zero, the runtime per batch is stable from the start.

	% Table generated by Excel2LaTeX from sheet 'All_class_single'
	\begin{table}[htbp]
		\centering
		\caption{Relative and absolute (seconds, in brackets) single-core average batch runtimes of \textsc{XVSelect}, \mbox{\textsc{XVSelect+RC}}, \textsc{Custom}, \textsc{Custom+RC} strategies on classification dataset \#28 (Power Italy) for different classification algorithms with $n=50$ and NB base learner.}
		\begin{tabular}{lllll}
			\textbf{Adaptive Strategy} & \textbf{bPL}   & \textbf{bBLAST} & \textbf{bLB}   & \textbf{bDWM} \\ \hline
			\textsc{Custom} & 1 (0.036) & 1 (0.004) & 1 (0.179) & 1 (0.056) \\
			\hline
			\textsc{XVSelect} (2 folds) & 2.687 (0.098) & 7.67 (0.033) & 3.594 (0.644) & 112.232 (6.246) \\
			\hline
			\textsc{XVSelect} (5 folds) & 4.896 (0.178) & 11.678 (0.05) & 8.901 (1.595) & 232.83 (12.957) \\
			\hline
			\textsc{XVSelect} (10 folds) & 8.797 (0.32) & 17.2 (0.074) & 17.622 (3.158) & 455.76 (25.363) \\
			\hline
			\textsc{Custom+RC} & 1.003 (0.036) & 6.692 (0.029) & 3.008 (0.539) & 35.105 (1.954) \\
			\hline
			\textsc{XVSelect+RC} (2 folds) & 2.676 (0.097) & 13.205 (0.057) & 6.062 (1.086) & 110.728 (6.162) \\
			\hline
			\textsc{XVSelect+RC} (5 folds) & 5.078 (0.184) & 14.912 (0.064) & 11.348 (2.033) & 224.421 (12.489) \\
			\hline
			\textsc{XVSelect+RC} (10 folds) & 9.047 (0.329) & 17.961 (0.077) & 20.16 (3.613) & 432.138 (24.049) \\
			\hline
		\end{tabular}%
		\label{tab:runtimes}%
	\end{table}%
	
	\section{Discussion and Conclusions}
	\label{section:conclusions}
	
	The core aim of this paper was to explore the issue of automating the adaptation of predictive algorithms, which was found to be a rather overlooked direction in otherwise popular area of automated machine learning. In our research, we have addressed this by utilising a simple, yet powerful adaptation framework, which separates adaptation from prediction, defines adaptive mechanisms and adaptive strategies, as well as allows the use of retrospective model correction. This adaptation framework enables the development of generic automated adaptation strategies, which can be deployed on any set of adaptive mechanisms, thus facilitating the automation of predictive algorithms' adaptation.
	
	We have used several automated adaptation strategies, based on cross-validation on the current batch and retrospectively reverting the model to the oracle state after obtaining the most recent batch of data. We postulate that the recently seen data is likely to be more related to the incoming data, therefore these strategies tend to steer the adaptation of the predictive model to achieve better results on the most recent available data. 
	
	To confirm our assumptions, we have empirically investigated the merit of automated adaptation strategies \textsc{XVSelect} and \textsc{XVSelect+RC}. For this purpose we have conducted experiments on 10 real and 26 synthetic datasets, exhibiting various levels of adaptation need. 
	
	The results are promising, as for the majority of these datasets, the proposed automated approaches were able to demonstrate comparable or better performance to those of specifically designed custom algorithms and the repeated deployment of any single adaptive mechanism. However, it is not the goal of this paper  to replace existing custom strategies with the proposed ones. We rather see the benefit of the proposed strategies in their applicability to all algorithms with multiple adaptive mechanisms, so that the designer of the algorithm does not need to spend time and effort to develop a custom adaptive strategy. We have analysed the cases where proposed strategies performed relatively poorly. It is postulated that the reasons for these cases were: a) lack of change/need for adaptation; b) insufficient data in a batch; and c) relatively simple datasets, all of which have trivial solutions. We have also identified that the choice of algorithm and base learner can affect the performance of proposed strategies.
	
	A benefit of the proposed generic automated adaptation strategies is that they can help designers of machine learning solutions save time by not having to devise a custom adaptive strategy. \textsc{XVSelect} and \textsc{XVSelect+RC} are generally parameter-free, except for the number of cross validation folds, choosing which is trivial. 
	
	Naturally, the described strategies come at some cost in performance. This cost varies between different algorithms and is dependent on the number of AMs and other factors, such as number of experts. The runtimes can be reduced by the parallelisation of cross-validatory selection and retrospective correction. It is also conceivable for throughput requirements to be lower for batch learning scenario, as the data is passed to the model only after the whole batch is accumulated.
	
	\section{Future Work}
	\label{section:future}
	This research has focused on batch scenario. Adapting the introduced automated adaptive strategies for incremental learning scenario remains a future research question. In that case a lack of batches would for example pose a question of data selection for cross validation. This could be addressed using data windows of static or dynamically changing size. Using an alternative to cross validation can be another solution. Another useful scope of research is focusing on a semi-supervised scenario, where true values or labels are not always available. This is relevant for many applications, amongst them in the process industry. 
	
	A dimension which may require more attention is further improvement of the runtime performance of the proposed approaches. An obvious first step in this direction is discarding the less useful, such as ``do nothing'', AMs.
	
	Further research directions include theoretical analysis of this direction of research, where relevant expert/bandit strategies may be useful, as well as the experiments with other ML tasks such as time series prediction, clustering and recommender systems. Finally, as we have observed some discrepancies in performance of the proposed approaches across algorithms/datasets/base learners, a natural research direction is to investigate the reasons for these discrepancies. This would also include experimentation with different base learners. 
	
	In general, there is a rising tendency of modular systems for construction of machine learning solutions, where adaptive mechanisms are considered as separate entities, along with pre-processing and predictive techniques. One of the features of such systems is easy, and often automated plug-and-play machine learning  \citep{Kadlec2009,kedziora2020autonoml}.  Generic automated adaptive strategies introduced in this paper further contribute towards this automation. 
	
	\begin{acknowledgements}
		We are grateful to anonymous reviewers for their valuable input. We also would like to thank Evonik Industries AG for the provided datasets. Part of the used Matlab code originates from Petr Kadlec and Ratko Grbi\'{c}. Tomasz Maszczyk has provided helpful coding advice. Thanks to R\'{o}man Arango for sharing his statistical wisdom.
	\end{acknowledgements}

	\clearpage
	
	\appendix
	\section{Supplementary material.}
	\label{appendixa}
	
	\begin{table}[H]
		\caption{Regression datasets with $\boldsymbol{N}$ instances and  $\boldsymbol{M}$ features.}
		\centering
		\begin{tabularx}{\linewidth}{lp{1.05cm}rlX}
			\hline
			\textbf{\#} & \textbf{Name} & $\boldsymbol{N}$ & $\boldsymbol{M}$  & \textbf{Description} \\
			\hline
			1 & Catalyst activation & 5867  & 12    & Highly volatile simulation (real conditions based) of catalyst activation in a multi-tube reactor.  Task is the prediction of catalyst activity while inputs are flows, concentrations and temperatures \citep{Strackeljan2006}. \\ \hline
			2 & Thermal oxidiser & 2820  & 36    & Prediction of $NO_x$ exhaust gas concentration during an industrial process, moderately volatile. Input features include  concentrations, flows, pressures and temperatures \citep{Kadlec2009}. \\ \hline
			3 & Industrial drier & 1219  & 16    & Prediction of residual humidity of the process product, relatively stable. Input features include temperatures, pressures and humidities \citep{Kadlec2009}. \\ \hline
			4 & Debutaniser column & 2394  & 7     & Prediction of butane concentration at the output of the column. Input features are temperatures, pressures and flows \citep{Fortuna2005}. \\ \hline
			5 & Sulfur recovery & 10081 & 6     & Prediction of $SO_2$ in the output of sulfur recovery unit. Input features are gas and air flow measurements \citep{Fortuna2003}. \\ \hline
		\end{tabularx}%
		
		\label{table:regressiondatareal}%
	\end{table}%

\begin{table}[H]
	\caption{Real world classification datasets with $\boldsymbol{N}$ instances,  $\boldsymbol{M}$ features and $\boldsymbol{C}$ classes.}
	\centering
	\begin{tabularx}{\linewidth}{rp{1.35cm}rrrX}
		\hline
		\textbf{\#} & \textbf{Name} &  $\boldsymbol{N}$ & $\boldsymbol{M}$ & $\boldsymbol{C}$& \textbf{Brief description} \\
		\hline
		27 & Australian electricity prices (Elec2) & 27887  & 6 & 2 & Widely used concept drift benchmark dataset thought to have seasonal and other changes as well as noise.  Task is the prediction of whether electricity price rises or falls while inputs are days of the week, times of the day and electricity demands \citep{Harries1999}. \\ \hline
		28 & Power Italy & 4489  & 2 & 4 & The task is prediction of hour of the day (03:00, 10:00, 17:00 and 21:00) based on supplied and transferred power measured in Italy. \citep{Zhu2010,UCRArchive}. \\ \hline
		29 & Contraceptive & 4419  & 9  & 3  & Contraceptive dataset from UCI repository \citep{Newman1998} with artificially added drift \citep{Minku2010}. \\ \hline
		30 & Iris & 450  & 4  & 4   & Iris dataset 
		\citep{Anderson1936,Fisher1936} with artificially added drift \citep{Minku2010}. \\ \hline
		31 & Yeast & 5928 & 8 & 10    & Contraceptive dataset from UCI repository \citep{Newman1998} with artificially added drift \citep{Minku2010}. \\ \hline
	\end{tabularx}%
	\label{table:classificationdatareal}
\end{table}%

	\begin{table}[H]
		\caption[Synthetic classification datasets used in experiments.]{Synthetic classification datasets used in experiments, with $\boldsymbol{N}$ instances and $\boldsymbol{C}$ classes, from \citep{Bakirov2013}. Column ``Drift" specifies number of drifts/changes in data, the percentage of change in the decision boundary and its type. All datasets have 2 input features.}
		\begin{tabularx}{\linewidth}{llllXXX}
			\hline\noalign{\smallskip}
			\textbf{\#} & \textbf{Data type} & $\boldsymbol{N}$ & $\boldsymbol{C}$& \textbf{Drift} & \textbf{Noise/overlap}\\ 
			\noalign{\smallskip}
			\hline
			\noalign{\smallskip}
			1 & Hyperplane & 600 & 2 & 2x50\% rotation & None \\ \hline
			2 & Hyperplane & 600 & 2 & 2x50\% rotation & 10\% uniform noise \\ \hline
			3 & Hyperplane & 600 & 2 & 9x11.11\% rotation & None \\ \hline
			4 & Hyperplane & 600 & 2 & 9x11.11\% rotation & 10\% uniform noise \\ \hline
			5 & Hyperplane & 640 & 2 & 15x6.67\% rotation & None \\ \hline
			6 & Hyperplane & 640 & 2 & 15x6.67\% rotation & 10\% uniform noise \\	 \hline
			7 & Hyperplane & 1500 & 4 & 2x50\% rotation & None  \\  \hline
			8 & Hyperplane & 1500 & 4 & 2x50\% rotation & 10\% uniform noise \\  \hline
			9 & Gaussian & 1155 & 2 & 4x50\% switching & 0-50\% overlap \\  \hline
			10 & Gaussian & 1155 & 2 & 10x20\% switching & 0-50\% overlap \\  \hline
			11 & Gaussian & 1155 & 2 & 20x10\% switching & 0-50\% overlap \\  \hline
			12 & Gaussian & 2805 & 2 & 4x49.87\% passing & 0.21-49.97\% overlap \\  \hline
			13 & Gaussian & 2805 & 2 & 6x27.34\% passing & 0.21-49.97\% overlap \\ \hline
			14 & Gaussian & 2805 & 2 & 32x9.87\% passing & 0.21-49.97\% overlap \\   \hline
			15 & Gaussian & 945 & 2 & 4x52.05\% move & 0.04\% overlap \\  \hline
			16 & Gaussian & 945 & 2 & 4x52.05\% move & 10.39\% overlap \\  \hline
			17 & Gaussian & 945 & 2 & 8x27.63\% move & 0.04\% overlap\\  \hline
			18 & Gaussian & 945 & 2 & 8x27.63\% move & 10.39\% overlap \\ \hline
			19 & Gaussian & 945 & 2 & 20x11.25\% move & 0.04\% overlap \\  \hline
			20 & Gaussian & 945 & 2 & 20x11.25\% move & 10.39\% overlap \\   \hline
			21 & Gaussian & 1890 & 4 & 4x52.05\% move & 0.013\% overlap \\  \hline
			22 & Gaussian & 1890 & 4 & 4x52.05\% move & 10.24\% overlap \\  \hline
			23 & Gaussian & 1890 & 4 & 8x27.63\% move & 0.013\% overlap\\  \hline
			24 & Gaussian & 1890 & 4 & 8x27.63\% move & 10.24\% overlap \\ \hline
			25 & Gaussian & 1890 & 4 & 20x11.25\% move & 0.013\% overlap \\  \hline
			26 & Gaussian & 1890 & 4 & 20x11.25\% move & 10.24\% overlap \\  \hline 
		\end{tabularx}
		\label{table:classificationdatasynth}
	\end{table}

	\begin{figure}[H]
	\centering
	\includegraphics[width=0.9\linewidth]{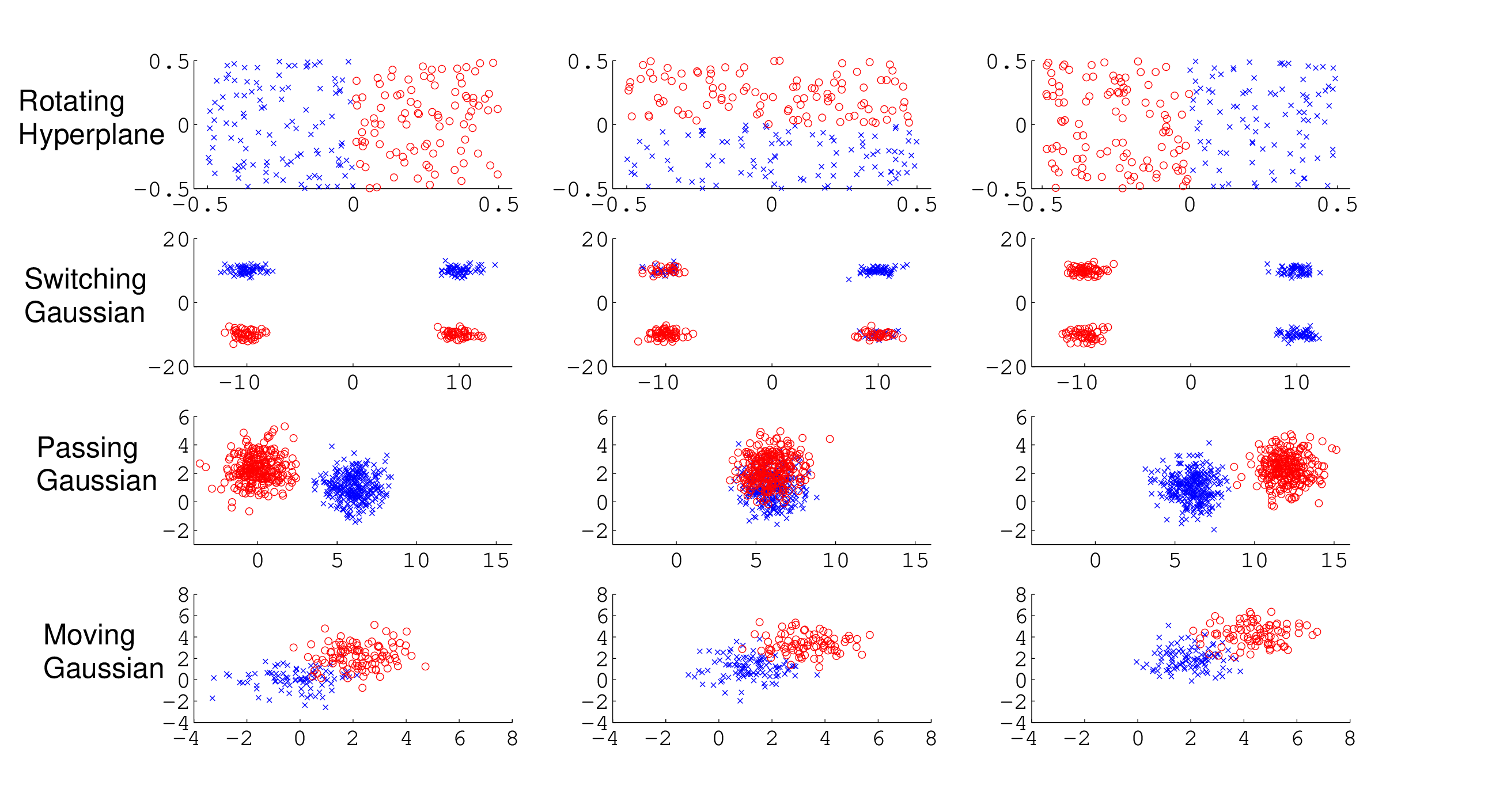}
	\caption{Synthetic datasets visualisation \citep{Bakirov2013}.}
	\label{fig:synthdata}
	\end{figure}

	% Table generated by Excel2LaTeX from sheet 'JournPapSimple'
	\begin{table}[H]
		\caption[SABLE hyperparameters for different datasets.]{SABLE hyperparameters for different datasets with batch size $n$, update weights of descriptors $\delta_0,\delta_1$, RPLS forgetting factor $\lambda$, kernel width for descriptor construction $\sigma$, $L$ RPLS latent variables and $K$ batches}
		\centering
		\begin{tabularx}{\columnwidth}{XXXrrrr}
			\hline
			\textbf{Dataset} & $n$ & $K$ &  $\delta_0,\delta_1$ & $\lambda$ & $\sigma$ & $L$\\
			\hline
			\textbf{Catalyst} & 50 & 117  & 0, 1  & 0.5   & 1 & 12 \\
			\textbf{Catalyst} & 100 & 59  & 0, 1  & 0.25  & 1 & 12 \\
			\textbf{Catalyst} & 200 & 30  & 0, 1  & 0.5   & 1 & 12 \\
			\textbf{Oxidizer} & 50 & 47 & 0.25, 0.75 & 0.5   & 1 & 3\\
			\textbf{Oxidizer} & 100 & 29  & 0, 1  & 0.25  & 0.01 & 3\\
			\textbf{Oxidizer} & 200 & 15  & 0, 1  & 0.25  & 0.01 & 3\\
			\textbf{Drier} & 50 & 25  & 0, 1  & 0.25  & 0.01 & 16\\
			\textbf{Drier} & 100 & 13  & 0, 1  & 0.5   & 0.1 & 16\\
			\textbf{Drier} & 200 & 7 & 0, 1  & 0.25  & 0.01 & 16\\
			\textbf{Debutaniser} & 50 & 47  & 0.25, 0.75  & 0.5 & 1 & 6 \\
			\textbf{Debutaniser} & 100 & 23  & 0.25, 0.75  & 0.25 & 1 & 6 \\
			\textbf{Debutaniser} & 200 & 11  & 0, 1  & 0.5   & 1 & 6\\
			\textbf{Sulfur} & 50 & 201 & 0.25, 0.75  & 0.5  & 1 & 7\\
			\textbf{Sulfur} & 100 & 100 & 0, 1  & 0.5   & 0.1 & 7\\
			\textbf{Sulfur} & 200  & 50 & 0, 1 & 0.5   & 0.1 & 7\\
			\hline
		\end{tabularx}%
		
		\label{tab:settings}%
	\end{table}%
	
	
	
	\FloatBarrier %clearpage to make sure that references appear after the floats
	\bibliographystyle{spbasic}
	\bibliography{library}

\begin{thebibliography}{90}
\providecommand{\natexlab}[1]{#1}
\providecommand{\url}[1]{{#1}}
\providecommand{\urlprefix}{URL }
\expandafter\ifx\csname urlstyle\endcsname\relax
  \providecommand{\doi}[1]{DOI~\discretionary{}{}{}#1}\else
  \providecommand{\doi}{DOI~\discretionary{}{}{}\begingroup
  \urlstyle{rm}\Url}\fi
\providecommand{\eprint}[2][]{\url{#2}}

\bibitem[{Alcob{\'{e}}(2004)}]{Alcobe}
Alcob{\'{e}} JR (2004) {Incremental Hill-Climbing Search Applied to Bayesian
  Network Structure Learning}. In: Proceedings of the Eighth European
  Conference on Principles and Practice of Knowledge Discovery in Databases,
  Volume 3202 of Lecture Notes in Computer Science. Springer Verlag

\bibitem[{Alippi et~al.(2012)Alippi, Boracchi, and Roveri}]{Alippi2012}
Alippi C, Boracchi G, Roveri M (2012) {Just-in-time ensemble of classifiers}.
  In: The 2012 International Joint Conference on Neural Networks (IJCNN), IEEE,
  pp 1--8

\bibitem[{Anderson(1936)}]{Anderson1936}
Anderson E (1936) {The Species Problem in Iris}. Annals of the Missouri
  Botanical Garden 23(3):457

\bibitem[{Ba and Frey(2013)}]{Ba2013}
Ba J, Frey B (2013) {Adaptive dropout for training deep neural networks}. In:
  NIPS'13 Proceedings of the 26th International Conference on Neural
  Information Processing Systems, pp 3084--3092

\bibitem[{Bach and Maloof(2010)}]{Bach2010}
Bach S, Maloof M (2010) {A bayesian approach to concept drift}. In: Advances in
  Neural Information, pp 127--135

\bibitem[{Bakirov(2017)}]{Bakirov2017a}
Bakirov R (2017) {Multiple adaptive mechanisms for predictive models on
  streaming data}. PhD thesis, Bournemouth University

\bibitem[{Bakirov and Gabrys(2013)}]{Bakirov2013}
Bakirov R, Gabrys B (2013) {Investigation of Expert Addition Criteria for
  Dynamically Changing Online Ensemble Classifiers with Multiple Adaptive
  Mechanisms}. In: Papadopoulos H, Andreou A, Iliadis L, Maglogiannis I (eds)
  Artificial Intelligence Applications and Innovations, vol 412, pp 646--656

\bibitem[{Bakirov et~al.(2015)Bakirov, Gabrys, and Fay}]{Bakirov2015}
Bakirov R, Gabrys B, Fay D (2015) {On sequences of different adaptive
  mechanisms in non-stationary regression problems}. In: 2015 International
  Joint Conference on Neural Networks (IJCNN), pp 1--8

\bibitem[{Bakirov et~al.(2016)Bakirov, Gabrys, and Fay}]{Bakirov2016}
Bakirov R, Gabrys B, Fay D (2016) {Augmenting adaptation with retrospective
  model correction for non-stationary regression problems}. In: 2016
  International Joint Conference on Neural Networks (IJCNN), IEEE, pp 771--779

\bibitem[{Bakirov et~al.(2017)Bakirov, Gabrys, and Fay}]{Bakirov2017}
Bakirov R, Gabrys B, Fay D (2017) {Multiple adaptive mechanisms for data-driven
  soft sensors}. Computers {\&} Chemical Engineering 96:42--54

\bibitem[{Bifet and Gavald{\`{a}}(2007)}]{Bifet2007}
Bifet A, Gavald{\`{a}} R (2007) {Learning from time-changing data with adaptive
  windowing}. SIAM International Conference on Data Mining 7:443--448

\bibitem[{Bifet et~al.(2009)Bifet, Holmes, Gavald{\`{a}}, Pfahringer, and
  Kirkby}]{Bifet2009}
Bifet A, Holmes G, Gavald{\`{a}} R, Pfahringer B, Kirkby R (2009) {New Ensemble
  Methods For Evolving Data Streams}. Proceedings of the 15th ACM SIGKDD
  international conference on Knowledge discovery and data mining - KDD '09 pp
  139--147

\bibitem[{Bifet et~al.(2010{\natexlab{a}})Bifet, Holmes, Kirkby, and
  Pfahringer}]{Bifet2010a}
Bifet A, Holmes G, Kirkby R, Pfahringer B (2010{\natexlab{a}}) {MOA: Massive
  Online Analysis}. Journal of Machine Learning Research 11(52):1601--1604

\bibitem[{Bifet et~al.(2010{\natexlab{b}})Bifet, Holmes, and
  Pfahringer}]{Bifet2010}
Bifet A, Holmes G, Pfahringer B (2010{\natexlab{b}}) {Leveraging bagging for
  evolving data streams}. In: Lecture Notes in Computer Science (including
  subseries Lecture Notes in Artificial Intelligence and Lecture Notes in
  Bioinformatics), vol 6321 LNAI, pp 135--150

\bibitem[{Cardillo(2009)}]{Cardillo2009}
Cardillo G (2009) {MYFRIEDMAN: Friedman test for non parametric two way
  ANalysis Of VAriance}.
  \urlprefix\url{https://www.mathworks.com/matlabcentral/fileexchange/25882-myfriedman}

\bibitem[{Carnein et~al.(2020)Carnein, Trautmann, Bifet, and
  Pfahringer}]{Carnein2020}
Carnein M, Trautmann H, Bifet A, Pfahringer B (2020) {Towards automated
  configuration of stream clustering algorithms}. In: Communications in
  Computer and Information Science, Springer, vol 1167 CCIS, pp 137--143

\bibitem[{Carpenter et~al.(1991)Carpenter, Grossberg, and
  Reynolds}]{Carpenter1991}
Carpenter G, Grossberg S, Reynolds J (1991) {ARTMAP: Supervised real-time
  learning and classification of nonstationary data by a self-organizing neural
  network}. Neural networks 4:565--588

\bibitem[{Castillo and Gama(2006)}]{Castillo2006}
Castillo G, Gama J (2006) {An Adaptive Prequential Learning Framework for
  Bayesian Network Classifiers}. In: F{\"{u}}rnkranz J, Scheffer T,
  Spiliopoulou M (eds) Knowledge Discovery in Databases: PKDD 2006, Springer
  Berlin Heidelberg, Berlin, Heidelberg, Lecture Notes in Computer Science, vol
  4213, pp 67--78

\bibitem[{Celik and Vanschoren(2020)}]{Celik2020}
Celik B, Vanschoren J (2020) {Adaptation Strategies for Automated Machine
  Learning on Evolving Data. arXiv pre-print},
  \urlprefix\url{http://arxiv.org/abs/2006.06480}

\bibitem[{Chen et~al.(2015)Chen, Keogh, Hu, Begum, Bagnall, Mueen, and
  Batista}]{UCRArchive}
Chen Y, Keogh E, Hu B, Begum N, Bagnall A, Mueen A, Batista G (2015) {The UCR
  Time Series Classification Archive}

\bibitem[{Cinar et~al.(2003)Cinar, Parulekar, Undey, and Birol}]{Cinar2003}
Cinar A, Parulekar SJ, Undey C, Birol G (2003) {Batch Fermentation: Modeling:
  Monitoring, and Control}. CRC Press

\bibitem[{Dawid(1984)}]{Dawid1984}
Dawid AP (1984) {Present Position and Potential Developments: Some Personal
  Views: Statistical Theory: The Prequential Approach}. Journal of the Royal
  Statistical Society Series A (General) 147(2):278

\bibitem[{Dem{\v{s}}ar(2006)}]{Demsar2006}
Dem{\v{s}}ar J (2006) {Statistical Comparisons of Classifiers over Multiple
  Data Sets}. Journal of Machine Learning Research 7(Jan):1--30

\bibitem[{Domingos and Hulten(2000)}]{Domingos2000}
Domingos P, Hulten G (2000) {Mining high-speed data streams}. Proceedings of
  the sixth ACM SIGKDD international conference on Knowledge discovery and data
  mining - KDD '00 pp 71--80

\bibitem[{DrawNemenyi(2019)}]{drawnemenyi}
DrawNemenyi (2019) {drawNemenyi}.
  \urlprefix\url{https://github.com/sepehrband/drawNemenyi}

\bibitem[{Duin et~al.(2007)Duin, Juszczak, Paclik, Pekalska, de~Ridder, Tax,
  and Verzakov}]{PRTools}
Duin RPW, Juszczak P, Paclik P, Pekalska E, de~Ridder D, Tax DMJ, Verzakov S
  (2007) {PRTools4.1, A Matlab Toolbox for Pattern Recognition”}

\bibitem[{Elwell and Polikar(2011)}]{Elwell2011}
Elwell R, Polikar R (2011) {Incremental learning of concept drift in
  nonstationary environments.} IEEE transactions on neural networks / a
  publication of the IEEE Neural Networks Council 22(10):1517--31

\bibitem[{Fern and Givan(2000)}]{Fern2000}
Fern A, Givan R (2000) {Dynamic feature selection for hardware prediction}.
  Tech. rep., Purdue University

\bibitem[{Feurer et~al.(2015)Feurer, Klein, Eggensperger, Springenberg, Blum,
  and Hutter}]{Feurer2015}
Feurer M, Klein A, Eggensperger K, Springenberg J, Blum M, Hutter F (2015)
  {Efficient and Robust Automated Machine Learning}. In: Advances in Neural
  Information Processing Systems 28 (NIPS 2015), pp 2962--2970

\bibitem[{Fisher(1936)}]{Fisher1936}
Fisher RA (1936) {The Use of Multiple Measurements in Taxonomic Problems}.
  Annals of Eugenics 7(2):179--188

\bibitem[{Fortuna et~al.(2003)Fortuna, Rizzo, Sinatra, and
  Xibilia}]{Fortuna2003}
Fortuna L, Rizzo A, Sinatra M, Xibilia M (2003) {Soft analyzers for a sulfur
  recovery unit}. Control Engineering Practice 11(12):1491--1500

\bibitem[{Fortuna et~al.(2005)Fortuna, Graziani, and Xibilia}]{Fortuna2005}
Fortuna L, Graziani S, Xibilia M (2005) {Soft sensors for product quality
  monitoring in debutanizer distillation columns}. Control Engineering Practice
  13(4):499--508

\bibitem[{Friedman and Goldszmidt(1997)}]{Friedman1997}
Friedman N, Goldszmidt M (1997) {Sequential update of Bayesian network
  structure}. In: Proceedings of the Thirteenth conference on Uncertainty in
  artificial intelligence, pp 165--174

\bibitem[{Gabrys(2004)}]{Gabrys2004}
Gabrys B (2004) {Learning hybrid neuro-fuzzy classifier models from data: to
  combine or not to combine?} Fuzzy Sets and Systems 147(1):39--56

\bibitem[{Gabrys and Bargiela(1999)}]{Gabrys1999}
Gabrys B, Bargiela A (1999) {Neural Networks Based Decision Support in Presence
  of Uncertainties}. Journal of Water Resources Planning and Management
  125(5):272--280

\bibitem[{Gabrys and Ruta(2006)}]{Gabrys2006}
Gabrys B, Ruta D (2006) {Genetic algorithms in classifier fusion}. Applied Soft
  Computing 6(4):337--347

\bibitem[{{Gomes Soares} and Ara{\'{u}}jo(2015)}]{GomesSoares2015}
{Gomes Soares} S, Ara{\'{u}}jo R (2015) {An on-line weighted ensemble of
  regressor models to handle concept drifts}. Engineering Applications of
  Artificial Intelligence 37:392--406

\bibitem[{Hall et~al.(2009)Hall, Frank, Holmes, Pfahringer, Reutemann, and
  Witten}]{Hall2009}
Hall M, Frank E, Holmes G, Pfahringer B, Reutemann P, Witten IH (2009) {The
  WEKA data mining software: an update}. ACM SIGKDD Explorations Newsletter
  11(1):10

\bibitem[{Harries(1999)}]{Harries1999}
Harries M (1999) {Splice-2 comparative evaluation: Electricity pricing.
  Technical report. The University of South Wales}. Tech. rep., The University
  of South Wales

\bibitem[{Hazan and Seshadhri(2009)}]{Hazan2009}
Hazan E, Seshadhri C (2009) {Efficient learning algorithms for changing
  environments}. In: ICML '09 Proceedings of the 26th Annual International
  Conference on Machine Learning, pp 393--400

\bibitem[{Herbster and Warmuth(1998)}]{Herbster1998}
Herbster M, Warmuth M (1998) {Tracking the best expert}. Machine Learning
  29:1--29

\bibitem[{Hulten et~al.(2001)Hulten, Spencer, and Domingos}]{Hulten2001}
Hulten G, Spencer L, Domingos P (2001) {Mining time-changing data streams}. In:
  Proceedings of the seventh ACM SIGKDD international conference on Knowledge
  discovery and data mining - KDD '01, ACM Press, New York, New York, USA, pp
  97--106

\bibitem[{Hutter et~al.(2011)Hutter, Hoos, and Leyton-Brown}]{Hutter2011}
Hutter F, Hoos HH, Leyton-Brown K (2011) {Sequential Model-Based Optimization
  for General Algorithm Configuration}. In: LION'05 Proceedings of the 5th
  international conference on Learning and Intelligent Optimization, Springer,
  Berlin, Heidelberg, pp 507--523

\bibitem[{Ikonomovska et~al.(2010)Ikonomovska, Gama, and
  D{\v{z}}eroski}]{Ikonomovska2010}
Ikonomovska E, Gama J, D{\v{z}}eroski S (2010) {Learning model trees from
  evolving data streams}. Data Mining and Knowledge Discovery 23(1):128--168

\bibitem[{Jang et~al.(1997)Jang, Sun, and Mizutani}]{Jang1997}
Jang JSR, Sun CT, Mizutani E (1997) {Neuro-Fuzzy and Soft Computing: A
  Computational Approach to Learning and Machine Intelligence}. Prentice Hall

\bibitem[{{Joe Qin}(1998)}]{JoeQin1998}
{Joe Qin} S (1998) {Recursive PLS algorithms for adaptive data modeling}.
  Computers {\&} Chemical Engineering 22(4-5):503--514

\bibitem[{Kadlec and Gabrys(2009)}]{Kadlec2009}
Kadlec P, Gabrys B (2009) {Architecture for development of adaptive on-line
  prediction models}. Memetic Computing 1(4):241--269

\bibitem[{Kadlec and Gabrys(2010)}]{Kadlec2010}
Kadlec P, Gabrys B (2010) {Adaptive on-line prediction soft sensing without
  historical data}. In: The 2010 International Joint Conference on Neural
  Networks (IJCNN), IEEE, pp 1--8

\bibitem[{Kadlec and Gabrys(2011)}]{Kadlec2011a}
Kadlec P, Gabrys B (2011) {Local learning-based adaptive soft sensor for
  catalyst activation prediction}. AIChE Journal 57(5):1288--1301

\bibitem[{Kedziora et~al.(2020)Kedziora, Musial, and
  Gabrys}]{kedziora2020autonoml}
Kedziora DJ, Musial K, Gabrys B (2020) Autonoml: Towards an integrated
  framework for autonomous machine learning. \eprint{2012.12600}

\bibitem[{Klinkenberg(2004)}]{Klinkenberg2004}
Klinkenberg R (2004) {Learning drifting concepts: Example selection vs .
  example weighting}. Intelligent Data Analysis 8(3):281--300

\bibitem[{Klinkenberg and Joachims(2000)}]{Klinkenberg2000}
Klinkenberg R, Joachims T (2000) {Detecting concept drift with support vector
  machines}. In: Proceedings of the Seventeenth International Conference on
  Machine Learning (ICML), pp 487--494

\bibitem[{Kolter and Maloof(2007)}]{coltermaloof2007}
Kolter JZ, Maloof MA (2007) {Dynamic weighted majority: An ensemble method for
  drifting concepts}. The Journal of Machine Learning Research Volume
  8,:2755--2790

\bibitem[{Kotthoff et~al.(2017)Kotthoff, Thornton, Hoos, Hutter, and
  Leyton-Brown}]{Kotthoff2017}
Kotthoff L, Thornton C, Hoos HH, Hutter F, Leyton-Brown K (2017) {Auto-WEKA
  2.0: Automatic model selection and hyperparameter optimization in WEKA}.
  Journal of Machine Learning Research 18(25):1--5

\bibitem[{Kuncheva(2004)}]{Kuncheva2004a}
Kuncheva LI (2004) {Combining Pattern Classifiers: Methods and Algorithms}.
  Wiley-Blackwell

\bibitem[{Lemke and Gabrys(2010)}]{Lemke2010}
Lemke C, Gabrys B (2010) {Meta-learning for time series forecasting and
  forecast combination}. Neurocomputing 73(10):2006--2016

\bibitem[{Lemke et~al.(2009)Lemke, Riedel, and Gabrys}]{Lemke2009}
Lemke C, Riedel S, Gabrys B (2009) {Dynamic combination of forecasts generated
  by diversification procedures applied to forecasting of airline
  cancellations}. In: 2009 IEEE Symposium on Computational Intelligence for
  Financial Engineering, IEEE, pp 85--91

\bibitem[{Littlestone and Warmuth(1994)}]{Littlestone1994}
Littlestone N, Warmuth M (1994) {The Weighted Majority Algorithm}. Information
  and Computation 108(2):212--261

\bibitem[{Lloyd et~al.(2014)Lloyd, Duvenaud, Grosse, Tenenbaum, and
  Ghahramani}]{Lloyd2014}
Lloyd JR, Duvenaud D, Grosse R, Tenenbaum JB, Ghahramani Z (2014) {Automatic
  construction and natural-language description of nonparametric regression
  models}. In: Proceedings of the Twenty-Eighth AAAI Conference on Artificial
  Intelligence, AAAI Press, pp 1242--1250

\bibitem[{Madrid et~al.(2019)Madrid, Escalante, Morales, Tu, Yu, Sun-Hosoya,
  Guyon, and Sebag}]{Madrid2019}
Madrid JG, Escalante HJ, Morales EF, Tu WW, Yu Y, Sun-Hosoya L, Guyon I, Sebag
  M (2019) {Towards AutoML in the presence of Drift: first results},
  \eprint{1907.10772}

\bibitem[{{Mart{\'{i}}n Salvador} et~al.(2016){Mart{\'{i}}n Salvador}, Budka,
  and Gabrys}]{MartinSalvador2016}
{Mart{\'{i}}n Salvador} M, Budka M, Gabrys B (2016) {Adapting Multicomponent
  Predictive Systems using Hybrid Adaptation Strategies with Auto-WEKA in
  Process Industry}. In: AutoML at ICML 2016, 2011, pp 1--8

\bibitem[{{Martin Salvador} et~al.(2019){Martin Salvador}, Budka, and
  Gabrys}]{MartinSalvador2019}
{Martin Salvador} M, Budka M, Gabrys B (2019) {Automatic Composition and
  Optimization of Multicomponent Predictive Systems With an Extended
  Auto-WEKA}. IEEE Transactions on Automation Science and Engineering
  16(2):946--959

\bibitem[{Minku et~al.(2010)Minku, White, and {Xin Yao}}]{Minku2010}
Minku L, White A, {Xin Yao} (2010) {The Impact of Diversity on Online Ensemble
  Learning in the Presence of Concept Drift}. IEEE Transactions on Knowledge
  and Data Engineering 22(5):730--742

\bibitem[{Mohr et~al.(2018)Mohr, Wever, and H{\"{u}}llermeier}]{Mohr2018}
Mohr F, Wever M, H{\"{u}}llermeier E (2018) {ML-Plan: Automated machine
  learning via hierarchical planning}. Machine Learning 107(8-10):1495--1515

\bibitem[{Montiel et~al.(2018)Montiel, Read, Bifet, and Kegl}]{Montiel2018}
Montiel J, Read J, Bifet A, Kegl B (2018) {Scikit-Multiflow: A Multi-output
  Streaming Framework}. Journal of Machine Learning Research 19(72):1--5

\bibitem[{Newman et~al.(1998)Newman, Hettich, Blake, and Merz}]{Newman1998}
Newman D, Hettich S, Blake C, Merz C (1998) {UCI repository of machine learning
  databases}

\bibitem[{Nguyen et~al.(2012)Nguyen, Woon, Ng, and
  Wan}]{Nguyen2012heterogeneous}
Nguyen H, Woon Y, Ng W, Wan L (2012) {Heterogeneous Ensemble for Feature Drifts
  in Data Streams}. In: Advances in Knowledge Discovery and Data Mining,
  Springer, pp 1--12

\bibitem[{Nguyen et~al.(2020)Nguyen, Maszczyk, Musial, Z{\"o}ller, and
  Gabrys}]{Nguyen2020}
Nguyen TD, Maszczyk T, Musial K, Z{\"o}ller MA, Gabrys B (2020) Avatar -
  machine learning pipeline evaluation using surrogate model. In: Berthold MR,
  Feelders A, Krempl G (eds) Advances in Intelligent Data Analysis XVIII,
  Springer International Publishing, Cham, pp 352--365

\bibitem[{Olson and Moore(2019)}]{Olson2019}
Olson RS, Moore JH (2019) {TPOT: A Tree-Based Pipeline Optimization Tool for
  Automating Machine Learning}. Springer, Cham, pp 151--160

\bibitem[{Oza and Russell(2001)}]{Oza2001}
Oza NC, Russell S (2001) {Online bagging and boosting}. In Artificial
  Intelligence and Statistics 2001 pp 105 -- 112

\bibitem[{van Rijn et~al.(2015)van Rijn, Holmes, Pfahringer, and
  Vanschoren}]{Rijn2015having}
van Rijn JN, Holmes G, Pfahringer B, Vanschoren J (2015) {Having a Blast:
  Meta-Learning and Heterogeneous Ensembles for Data Streams}. In: Data Mining
  (ICDM), 2015 IEEE International Conference on, IEEE, pp 1003--1008

\bibitem[{Rossi et~al.(2014)Rossi, {de Leon Ferreira}, Soares, and {De
  Souza}}]{Rossi2014meta}
Rossi ALD, {de Leon Ferreira} ACP, Soares C, {De Souza} BF (2014) {MetaStream:
  A meta-learning based method for periodic algorithm selection in
  time-changing data}. Neurocomputing 127:52--64

\bibitem[{Ruta et~al.(2011)Ruta, Gabrys, and Lemke}]{Ruta2011}
Ruta D, Gabrys B, Lemke C (2011) {A Generic Multilevel Architecture for Time
  Series Prediction}. IEEE Transactions on Knowledge and Data Engineering
  23(3):350--359

\bibitem[{Sahel et~al.(2007)Sahel, Bouchachia, Gabrys, and Rogers}]{Sahel2007}
Sahel Z, Bouchachia A, Gabrys B, Rogers P (2007) {Adaptive Mechanisms for
  Classification Problems with Drifting Data}. In: Proc. of the 11th
  International Conference on Knowledge-based Intelligent Engineering Systems
  (KES'2007), Springer, Berlin, Heidelberg, pp 419--426

\bibitem[{Schlimmer and Granger(1986)}]{Schlimmer1986}
Schlimmer JC, Granger RH (1986) {Beyond incremental processing: Tracking
  Concept Drift}. AAAI-86 Proceedings pp 502--507

\bibitem[{Schmidt and Lipson(2007)}]{Schmidt2007}
Schmidt M, Lipson H (2007) {Learning noise}. Proceedings of the 9th annual
  conference on Genetic and evolutionary computation - GECCO '07 pp 1680--1685

\bibitem[{Scholz and Klinkenberg(2007)}]{Scholz2007}
Scholz M, Klinkenberg R (2007) {Boosting Classifiers for Drifting Concepts}.
  Intelligent Data Analysis 11(1):1--40

\bibitem[{Souza and Ara{\'{u}}jo(2014)}]{Souza2014}
Souza F, Ara{\'{u}}jo R (2014) {Online Mixture of Univariate Linear Regression
  Models for Adaptive Soft Sensors}. In: IEEE Transactions on Industrial
  Informatics, vol~10, pp 937--945

\bibitem[{Stanley(2002)}]{Stanley2002}
Stanley KO (2002) {Evolving neural networks through augmenting topologies}.
  Evolutionary computation 10(2):99--127

\bibitem[{Strackeljan(2006)}]{Strackeljan2006}
Strackeljan J (2006) {NiSIS Competition 2006- Soft Sensor for the adaptive
  Catalyst Monitoring of a Multi–Tube Reactor}. Tech. rep., Universit{\"{a}}t
  Magdeburg

\bibitem[{Street and Kim(2001)}]{streetkim2001}
Street WN, Kim YS (2001) {A streaming ensemble algorithm (SEA) for large-scale
  classification}. Proceedings of the seventh ACM SIGKDD international
  conference on Knowledge discovery and data mining pp 377--382

\bibitem[{Vakil-Baghmisheh and
  Pave{\v{s}}i{\'{c}}(2003)}]{Vakil-Baghmisheh2003}
Vakil-Baghmisheh MT, Pave{\v{s}}i{\'{c}} N (2003) {A Fast Simplified Fuzzy
  ARTMAP Network}. Neural Processing Letters 17(3):273--316

\bibitem[{Veloso et~al.(2018)Veloso, Gama, and Malheiro}]{Veloso2018}
Veloso B, Gama J, Malheiro B (2018) {Self Hyper-Parameter Tuning for Data
  Streams}. In: Lecture Notes in Computer Science (including subseries Lecture
  Notes in Artificial Intelligence and Lecture Notes in Bioinformatics),
  Springer Verlag, vol 11198 LNAI, pp 241--255

\bibitem[{Wang et~al.(2003)Wang, Fan, Yu, and Han}]{Wang2003}
Wang H, Fan W, Yu PS, Han J (2003) {Mining concept-drifting data streams using
  ensemble classifiers}. In: Proceedings of the ninth ACM SIGKDD international
  conference on Knowledge discovery and data mining - KDD '03, ACM Press, New
  York, New York, USA, pp 226--235

\bibitem[{Wasserman(2000)}]{Wasserman2000}
Wasserman L (2000) {Bayesian Model Selection and Model Averaging}. Journal of
  Mathematical Psychology 44(1):92--107

\bibitem[{Widmer and Kubat(1996)}]{Widmer1996}
Widmer G, Kubat M (1996) {Learning in the presence of concept drift and hidden
  contexts}. Machine Learning 23(1):69--101

\bibitem[{Wilcoxon(1945)}]{Wilcoxon1945Individual}
Wilcoxon F (1945) {Individual Comparisons by Ranking Methods}. Biometrics
  Bulletin 1(6):80, \doi{10.2307/3001968}

\bibitem[{Zhu(2010)}]{Zhu2010}
Zhu X (2010) {Stream Data Mining Repository,
  http://www.cse.fau.edu/{\~{}}xqzhu/stream.html}

\bibitem[{Zliobaite(2011)}]{Zliobaite2011}
Zliobaite I (2011) {Combining Similarity in Time and Space for Training Set
  Formation under Concept Drift}. Intelligent Data Analysis 15(4):589--611

\bibitem[{Zliobaite and Kuncheva(2010)}]{Zliobaite2010}
Zliobaite I, Kuncheva LI (2010) {Theoretical Window Size for Classification in
  the Presence of Sudden Concept Drift}. Tech. rep., CS-TR-001-2010, Bangor
  University, UK

\end{thebibliography}
	
\end{document}